\documentclass[]{interact}

\usepackage{epstopdf}
\usepackage{subfigure}
\usepackage{graphicx}
\usepackage{float}
\usepackage{hyperref}
\usepackage{amsmath}
\usepackage{algorithm}
\usepackage{algpseudocode}

\usepackage{natbib}
\bibpunct[, ]{(}{)}{,}{a}{}{,}
\renewcommand\bibfont{\fontsize{10}{12}\selectfont}

\theoremstyle{plain}

\theoremstyle{definition}

\theoremstyle{remark}

\usepackage{verbatim}

\usepackage{rotating}

\usepackage{siunitx}
\begin{document}


\title{Integrating Spatiotemporal Features in LSTM for Spatially Informed COVID-19 Hospitalization Forecasting}

\author{
\name{Zhongying Wang\textsuperscript{a}\thanks{CONTACT Morteza Karimzadeh. Email: karimzadeh@colorado.edu}, Thoai D. Ngo\textsuperscript{b}, Hamidreza Zoraghein\textsuperscript{c}, Benjamin Lucas\textsuperscript{a}, and Morteza Karimzadeh\textsuperscript{a}}
\affil{\textsuperscript{a}Department of Geography, University of Colorado, Boulder, CO, USA }
{\textsuperscript{b}Heilbrunn Department of Population and Family Health, Columbia University Mailman School of Public Health, New York, NY, USA }
{\textsuperscript{c}Population Council, NY, USA }
}

\maketitle

\begin{abstract}

The COVID-19 pandemic's severe impact  highlighted the need for accurate and timely hospitalization forecasting to support effective healthcare planning. However, most forecasting models struggled, particularly during variant surges, when they were most needed. This study introduces a novel parallel-stream Long Short-Term Memory (LSTM) framework to forecast daily state-level incident hospitalizations in the United States. Our framework incorporates a spatiotemporal feature, Social Proximity to Hospitalizations (SPH), derived from Meta's Social Connectedness Index, to improve forecasts. SPH serves as a proxy for interstate population interaction, capturing transmission dynamics across space and time. Our architecture captures both short- and long-term temporal dependencies, and a multi-horizon ensembling strategy balances forecasting consistency and error. An evaluation against the COVID-19 Forecast Hub ensemble models during the Delta and Omicron surges reveals the superiority of our model. On average, our model surpasses the ensemble by 27, 42, 54, and 69 hospitalizations per state at the 7-, 14-, 21-, and 28-day horizons, respectively, during the Omicron surge. Data-ablation experiments confirm SPH's predictive power, highlighting its effectiveness in enhancing forecasting models. This research not only advances hospitalization forecasting but also underscores the significance of spatiotemporal features, such as SPH, in modeling the complex dynamics of infectious disease spread.

\end{abstract}


\textbf{\small KEY POLICY HIGHLIGHTS}
\begin{itemize}
    \item \small Deep learning can be used to more reliably forecast the spread of infectious diseases.
    \item \small Social media friendship data can help quantify interstate disease transmission.
    \item \small Spatial models that leverage multi-state data are more reliable for forecasting and policymaking. 
\end{itemize}

\section{Introduction} \label{introduction}

Since the first laboratory-confirmed case of COVID-19, an infectious disease caused by the SARS-CoV-2 virus,  in the United States (U.S.) in January 2020, there have been approximately 111 million total confirmed infections, 1.2 million deaths, and 7 million hospitalizations as of June 2023. Despite the end of the pandemic phase and declining mortality rates, COVID-19 remains a significant global health concern. According to the Centers for Disease Control and Prevention (CDC) COVID-19 Dashboard, the disease exhibited a peak weekly test positivity of 18\% in the U.S. in 2024. Although the recorded hospitalization rate of 4.8 per 10,000 population on August 10, 2024, may appear comparatively low, it underscores the continuing impact of the disease. According to communications received from the CDC, hospitals are mandated to report COVID-19 hospitalizations again starting in mid-November 2024, indicating the resurgence of the disease. The COVID-19 pandemic strained healthcare resources and overloaded hospitals, exacerbating the dramatic loss of human life. SARS-CoV-2 spreads rapidly, causing severe complications due to its high reproduction rate, the ability to spread via asymptomatic individuals, the prevalence of close-contact settings in densely populated areas, continual mutation into more transmissible variants, and the inconsistent application of preventive public health measures across the U.S. As a result, the demand for travel nurses surged during the pandemic, aligning with shifts in COVID-19 infection hotspots \citep{cole2021, longyear2020travel}. This was partially a geospatial problem related to the timely allocation of limited human and medical resources. Reliable geographic forecasting of COVID-19 hospital admissions could have alleviated this burden through policy-relevant decision-making and proactive allocation of resources in regional hotspots (i.e.\, areas of increased demand).

Accurately forecasting the geographic spread of COVID-19, particularly hospitalizations,  presented significant challenges. Early in the pandemic, the time between symptom onset and hospitalization varied from 3 to 10.4 days, averaging about 5 days \citep{faes2020time, galmiche2023sars}; this interval provided advance, albeit short, notice for hospitals to prepare for an upcoming influx of inpatients after observing a rise in test positivity. However, with the emergence of variants, such as Omicron (B.1.1.529), the lag between the surge in cases and hospitalizations shortened to around 3.6 days on average \citep{jansen2021investigation, song2022serial, grant2020prevalence, tanaka2022shorter, galmiche2023sars}, reducing the preparation window for healthcare providers and complicating forecasting efforts. Additionally, the rise of at-home testing contributed to significant underreporting of cases \citep{liu2021predicting, irons2021estimating}, rendering case reports unreliable and reinforcing the importance of forecasting hospitalizations directly, which are mandated to be reported by the Centers for Medicare and Medicaid Services (CMS). 

Lastly, unlike previous surges, hospitalization incidence largely decoupled from case incidence during the Omicron surge \citep{moore2022omicron}. Although the Omicron variant infected a much larger share of the population more rapidly than earlier variants, it did not result in a similar proportion of hospitalizations. This could be due to the immunity granted by vaccination, prior infections, and potentially the characteristics of the Omicron variant itself. Therefore, direct forecasting of hospitalizations, rather than a proportion of confirmed cases, is critical for intervention planning and resource allocation. Additionally, a model must balance learning from short- and long-term changes in hospitalization rates to perform optimally in the face of new variants. Moreover, to improve hospitalization forecasting, the disease spread dynamics should be taken into account. SARS-CoV-2 is an airborne virus; therefore, human interaction leads to disease diffusion in space and time, resulting in spatiotemporal hotspots. While infections in a county may be influenced by social connections, some of those result in hospitalizations, leading to additional regional spillover effects. This means that infection and hospitalization in one spatial jurisdiction can affect trends in connected jurisdictions \citep{wang2020sensitivity}, and successful incorporation of spatial dependencies and connectivity is a key factor in improving the quality of forecasts \citep{lucas2023spatiotemporal}. 

To address these challenges, we developed a novel hospitalization forecasting model based on deep learning with spatiotemporal features derived from social media friendship data. Our model leverages Meta’s Social Connectedness Index (SCI), which is derived from aggregated social media friendship counts across spatial units, to capture potential interstate transmission. Because our model operates at the spatial unit level, rather than the individual level, it mitigates concerns related to individual Facebook usage. The high correlation of regional COVID-19 caseloads with SCI values has been shown in previous studies \citep{kuchler2022jue}, and another study found that SCI is more predictive than SafeGraph’s cell-phone-based mobility data, albeit with a simpler tree-based ensemble model \citep{vahedi2021spatiotemporal}. Although we acknowledge that no dataset is fully representative, Facebook’s SCI data remains one of the most comprehensive resources available for measuring regional connectivity and its impact on disease spread.

The main contributions of this paper can be summarized as follows:

\begin{enumerate}
    \item Development of a novel parallel-stream architecture based on Long Short-Term Memory (LSTM) network to learn multi-scale temporal dependence for hospitalization forecasting.
    \item Development of novel interstate spatiotemporal features using social media-derived aggregate data to incorporate spatial dependence, improve performance, and reduce model variance. 
    \item Design of a multi-horizon ensembling strategy that balances between predictive performance and consistency in output sequences. 
    \item Integration of quantile loss for probabilistic forecasts for each state to characterize uncertainty and evaluation with the weighted interval score, which jointly penalizes dispersion, under- and over-prediction. 
\end{enumerate}

This paper also reports rigorous evaluations of our proposed spatiotemporal features, multi-horizon model performance, uncertainty, and compares our model against baseline models on data collected during the Omicron surge in the U.S.---the most challenging period for hospitalization forecasting during the pandemic, as well as the earlier Delta surge period. We also release open-source code, models, and data processing pipelines for reproducibility and reuse by the research community, as well as for application to other infectious diseases. 

\section{Background} \label{background}

 A comprehensive review of COVID-19 modeling identified approximately 22,000 publications related to the topic, spanning various modeling approaches, most notably compartmental and statistical models \citep{cao2022covid}.

The United States COVID-19 Forecast Hub (hereafter referred to as the Hub) is an online repository created by the U.S. Centers for Disease Control and Prevention (CDC) and an academic research lab at the University of Massachusetts Amherst. Launched in April 2020 \citep{cramer2022united}, the repository fulfills several crucial roles in monitoring COVID-19 in the U.S., including providing reported data on incident cases and deaths from Johns Hopkins University (JHU) Center for System Science and Engineering (CSSE) Dataset \citep{dong2020interactive} and incident hospitalizations from HealthData.gov\footnote{https://healthdata.gov/Hospital/COVID-19-Reported-Patient-Impact-and-Hospital-Capa/g62h-syeh}. The COVID-19 dataset collected during the pandemic is uniquely large with high spatial coverage, and well quality controlled, thanks to the stringent legal and regulatory mandates. This has provided an excellent benchmark dataset for evaluating the performance of new modeling approaches, one that we leverage here. Having a central point of data acquisition has proven to be a valuable resource for researchers. Additionally, the repository collects and publishes point and probabilistic forecasts of incident cases, deaths, hospitalizations, and cumulative deaths due to COVID-19 at national, state, and county levels in the U.S. submitted by various research groups employing different methodologies. These forecasts facilitate comparative analyses of model performance and the development of ensemble models by combining forecasts from multiple models (see section \ref{ForecastHub}).

Hospitalization forecasts submitted to the Hub vary in time horizon and resolution, ranging from 1 to 130 days ahead, with longer horizons being the focus of scenario modeling. In this study, we primarily focus on short-term (28-day) incident hospitalization forecasts---the Hub-standard focus for hospitalization forecasting. Although county-level forecasting can provide finer-grained insights, the CDC and the Hub standardized state-level hospitalization forecasts as their operational benchmark \citep{cramer2022united}. This decision was partly driven by the fact that county-level hospitalization data were not consistently available nationwide during the evaluation period. Moreover, focusing on state-level data ensures consistency and comparability across different research teams submitting forecasts to the Hub, while aligning with practical constraints of reporting and data reliability. Beyond point forecasts, the Hub mandates the submission of quantiles from 23 probability levels (including 0.01, 0.025, 0.05, 0.10, 0.15, 0.20, 0.25, 0.30, 0.35, 0.40, 0.45, 0.50, 0.55, 0.60, 0.70, 0.75, 0.80, 0.90, 0.95, 0.975, 0.99) for state-level incident hospitalization forecasts. 

\subsection{COVID Forecast Hub Ensemble} \label{ForecastHub}

The Hub uses the submitted forecasts from contributing teams to generate three distinct models: a baseline model for reference and two ensemble models, all of which are incorporated for comparison in our experiments. The \textit{COVIDhub-baseline} model operates as a persistence model for reference \citep{cramer2022united}, where the point forecast for any future time is equal to the most recently observed value. Ensemble methods, a meta-learning approach, combine predictions from multiple models to enhance predictive accuracy. Starting from the week of November 15, 2021, \textit{COVIDhub-4\_week\_ensemble} computes an equally weighted median of eligible forecasts for cases and hospitalizations. Although the ensemble may not consistently yield the most accurate forecast for individual dates, it demonstrates stable forecasting performance over time compared to other models within the Hub \citep{bracher2021evaluating}. By contrast, the forecasts generated by the \textit{COVIDhub-trained\_ensemble} are a weighted median of forecasts from the top-performing individual models, ranked based on the relative Weighted Interval Score (WIS) assessed on performance for previous forecast submissions \citep{bracher2021evaluating}.

\subsection{Compartmental Models} \label{compartmental}

Among the 17 models in the Hub that forecast incident hospitalizations, seven employ compartmental models, a traditional approach with a history of application in modeling infectious diseases. The general idea of compartmental models is to divide the population into compartments throughout the progression of the disease \citep{kendall1956deterministic}. The most basic model, SIR, divides people into Susceptible, Infected, and Recovered groups. The dynamics of an epidemic without considering the dynamics of birth and death can be expressed by the following ordinary differential equations: 

\begin{subequations} \label{compartmental_equation}
    \begin{align}
        \frac{d S}{dt} &= - \frac{\beta I S}{N} \\
        \frac{d I}{dt} &= \frac{\beta I S}{N} - \gamma I \\
        \frac{d R}{dt} &= \gamma I
    \end{align}
\end{subequations}

where $S$ is the susceptible population, $I$ is the number of the infected, $R$ is the number of the recovered, and $N$ represents the total population. $\beta$ represents the effective transmission rate and $\gamma$ represents the recovery rate. The reproduction number, denoted $R_{0}$, is calculated as the ratio of $\beta$ to $\gamma$. $R_{0}$ is an important descriptive parameter in epidemiology modeling and represents the average number of people who will be infected by a given infected person. Due to their interpretability, compartment models are often used to simulate different scenarios of an epidemic and evaluate the effect of different public health interventions (e.g., social distancing, lockdown, vaccine distribution). 

The more popular SEIR model extends the SIR model by introducing an Exposed compartment, which reflects the incubation period during which individuals are exposed to the pathogen but have not yet developed symptoms. This modification is encapsulated in the modified equations 1b and 1c, now referred to as Equations (2b) and (2c), respectively:

\begin{subequations} \label{seir_equation} 
    \setcounter{equation}{1}
    \begin{align} 
    \frac{d E}{dt} &= \frac{\beta I S}{N} - \sigma E \\ 
    \frac{d I}{dt} &= \sigma E - \gamma I
    \end{align} 
\end{subequations}

Here, $\sigma$ is the rate at which exposed individuals become infectious. 

A common criticism of basic compartmental models is that they assume spatiotemporal homogeneity of the spatial units, and oversimplify the complex disease processes \citep{ansumali2020very, getz2019adequacy}. The complex and dynamic patterns of COVID-19 transmission require updating the disease parameters with epidemiological expert supervision in compartmental models. SEIR models can be extended to better incorporate spatiotemporal heterogeneity and human mobility, enhancing their applicability to dynamic epidemic conditions. For instance, the JHUAPL-Bucky model integrates spatial mobility matrices and age-based contact matrices to capture interactions across regions and different age demographics \citep{cramer2022united}. Similarly, models like GLEAM developed by \citeauthor{balcan2010modeling} utilize mobility networks and spatial diffusion to model disease spread across geographic regions to account for local variations in population density and movement patterns. These models, including USC’s SI-kJalpha, which adjusts for heterogeneous infection rates due to variants and vaccination statuses \citep{srivastava2020fast}, attempt to capture complex realities of infectious disease transmission within compartmental frameworks. Moreover, recent developments, such as the Multistrain SEIR Model by \citeauthor{laaroussi2024optimal} and \citeauthor{seibel2024unifying}, integrate distributed vaccination policies and behavioral heterogeneity (which require additional data availability), respectively, incorporating spatial and human factors that influence disease dynamics. The JHUAPL-Bucky, GLEAM, and SI-kJalpha are included in the COVIDhub-4\_week\_ensemble, and due to their high performance, they are also included in the COVIDhub-trained\_ensemble, against which we compare our models in this paper. However, it is worth noting that deep learning-based models allow for more complex modeling through deeper networks with non-linear activation functions, with orders of magnitude more learnable parameters to capture heterogeneous dynamics.

\subsection{Statistical Models} \label{data-driven}

Due to the availability and high spatiotemporal resolution of COVID-19 incidence reports, statistical approaches have grown in popularity and demonstrated their performance in accurate forecasting during the COVID-19 pandemic \citep{reinhart2021open, rahimi2023review, vaughan2023exploration}. Whereas compartmental models make assumptions about population and epidemic characteristics, statistical approaches can learn complex patterns directly from observed data. However, these data-driven methods are highly dependent on data quality and usually have poor interpretability and scenario-building capability. 

As COVID-19 cases and hospitalizations are highly temporally correlated, many teams have chosen time-series forecasting methods within the family of statistical approaches. The Auto-Regressive Integrated Moving Average (ARIMA) model is a time-series forecasting method that makes future predictions by analyzing historical data. Some teams have applied such methods to forecast incident cases and hospitalizations, achieving satisfactory prediction accuracy \citep{benvenuto2020application, alabdulrazzaq2021accuracy}. Instead of using a single autoregressive method, CMU Delphi Group used a cumulative distribution function-space-averaged ensemble of three models, including a simple autoregressive model for point prediction, a quantile autoregressive model with additional case covariates, and a direction-stratified quantile autoregressive model \citep{reinhart2021open}. 

Deep learning-based approaches have shown great potential in epidemiological modeling in recent years. The GT-DeepCOVID model is a deep learning-based approach that learns the dependence of hospitalization and mortality rates from syndromic, demographic, mobility, and clinical data \citep{rodriguez2021deepcovid}. To address its poor interpretability, the model also integrates an explainability module to discover the impact of different variables. Long Short-Term Memory (LSTM) is an alternative deep learning-based approach that is more skillful at capturing long-distance dependencies in time-series data. LSTM and its variants have been applied in COVID-19 forecasting with COVID-19 time series data as input \citep{lucas2023spatiotemporal}.

\subsection{Social-Network-Derived Auxiliary Signals}

Real-time microblogging data have been investigated as auxiliary signals to improve epidemic forecasts. \citet{paul2014twitter} showed that geocoded influenza-related tweets reduced 1-week-ahead forecasting error by roughly 30\% and outperformed Google Flu Trends. A systematic review of 27 Twitter/Facebook surveillance studies has found consistent gains in timeliness over traditional surveillance systems \citep{alessa2018review}. Similar pipelines have been applied to COVID-19 to serve as predictors of case prevalence \citep{li2020data}. 

Despite these successes, social-media data have well-documented drawbacks. First, they suffer from demographic and geographic selection biases \citep{zhao2022biases}. Second, they are highly vulnerable to health-related misinformation, which can distort prevalence signals and even amplify epidemic spread \citep{ferrara2020misinformation, deverna2025modeling}. Finally, operational pipelines require continuous API access and natural language processing (NLP) workflows. These dependencies may introduce processing delays and break when platform policies change---constraints incompatible with the Hub's one-day window from data release to forecast submission. Nevertheless, Facebook’s extensive user base (with more than 2.5 billion active users globally and almost 200 million users in the U.S.) provides one of the most representative measures of broader social connectivity of spatial units, offering strong potential for improving epidemiological forecasting.

\subsection{Distinction from Existing Work}

Common measures of spatial autocorrelation assign weights to each region solely by geographic contiguity or distance, which only captures local spread and ignores non-local transmission pathways  \citep{anselin2005exploring, zhang2019epidemic}. Furthermore, they assume equal contributions from spatial neighbors, without quantifying the strength of the connection, to avoid introducing arbitrary structure into data. One of the distinctive highlights of our work is that it accounts for spatial spread by using Facebook’s Social Connectedness Index (SCI) to engineer spatiotemporal features. This index, which reflects the strength of relationships between regions in the social network space, has been shown to be associated with geographic proximity, historical ties, political boundaries, and other factors \citep{bailey2018social}. Social media-derived spatial connectivity features have demonstrated a strong positive association with population movement flows, and can serve as a proxy for average human movement measurements and population interaction \citep{li2021measuring}. This is also supported by earlier research \citep{kuchler2022jue}, showing a positive correlation between COVID-19 case clusters and SCI. 

While prior studies \citep{vahedi2021spatiotemporal, lucas2023spatiotemporal} have incorporated SCI-derived spatial features for county-level case forecasting, they did not present an ablation study, and they did not address the unique challenges of state-level hospitalization forecasting. 

Unlike previous COVID-19 case forecasting efforts, the COVID-19 Forecast Hub explicitly mandated daily, state-level, 28-day-ahead hospitalization forecasts, updated on a weekly schedule, throughout the pandemic. This is not a hypothetical exercise but an operational requirement with a core scientific challenge, as hospitalizations are considered a more reliable and policy-relevant target than cases---especially after the rise of at-home testing. Compared with county-level weekly case forecasts, this setup has fewer total data points (51 states) yet requires forecasting at a higher temporal resolution (daily rather than weekly). As a result, models must handle longer time-series with fewer spatial samples, making the learning task significantly more challenging than the discontinued case forecasting.

Our work is the first to leverage social media-derived connectivity for daily, state-level hospitalization forecasts within a deep learning framework. While a few studies incorporated SCI for county-level case predictions \citep{vahedi2021spatiotemporal, lucas2023spatiotemporal}, no prior work has harnessed these features for a 28-day operational hospitalization forecast. We address this gap by introducing our “SPH” feature to capture spatial spillover effects and integrating it into a parallel-stream LSTM architecture that learns both short- and long-term temporal dependencies. In detailed ablation experiments, we demonstrate how SPH significantly improves predictive accuracy and reduces variance---highlighting its added value in real-time policy contexts.

LSTMs, a type of recurrent neural network(RNN), excel in learning long-term temporal dependencies, making them ideal for sequence and time series data \citep{hochreiter1997long}. LSTM and its variants have been proven successful in applications of time-series epidemiological modeling \citep{venna2018novel}, especially COVID-19 Forecasting \citep{lucas2023spatiotemporal, chimmula2020time, shahid2020predictions}. Our model extends and improves this capability by incorporating parallel temporal lags in model inputs and a multi-horizon ensembling strategy to further improve time series forecasts.

\section{Materials and Methods}

This section delineates the intricacies of data processing, feature engineering, and model architecture. We explain how we enhanced our model's predictive capacity and robustness by leveraging features derived from Facebook’s Social Connectedness Index, integrating them into a novel parallel architecture of stacked Long Short-Term Memory (LSTM) networks. Additionally, we detail our novel multi-horizon ensemble strategy aimed at preserving both the consistency and accuracy of 28-day incident hospitalization forecasting.

\subsection{Data and Processing}

Our forecasting target, aligned with the Hub \citep{cramer2022united}, is the 28-day incident hospitalizations across 50 states of the U.S. plus Washington, D.C. Hence, on forecast date $t$, our predictions encompass values $y_{t+1}, y_{t+2}, \dots, y_{t+28}$ for each of the 51 locations. We designed an autoregressive model with three input feature time series: two temporal features derived from raw incident cases and hospitalizations, respectively, and a spatiotemporal feature representing the spatial spread of the virus (and its hospitalizations) across states, derived from Facebook's Social Connectedness Index. 

Incident case data were sourced from the U.S. COVID-19 Forecast Hub, and the incident cases at the state-level were calculated by summing incident counts across all sub-level (county-level) locations. State-level hospitalization data were retrieved from the Department of Health and Human Services website as the sum of confirmed daily adult and pediatric COVID-19 admissions.

We used social media connectivity to capture spatial dependence. The Social Connectedness Index (SCI) is a public dataset published by Meta and available in multiple spatial resolutions (e.g., counties, states, and countries) \citep{bailey2017measuring}. The index measures the strength of connectedness between two geographic areas as measured by Facebook friendship ties, providing a macro-level insight into social dynamics and disease spread potential rather than individual behaviors. Furthermore, Facebook’s extensive user base (with more than 2.5 billion active users globally and almost 200 million users in the U.S.), provides one of the most representative measures for broader social connectivity of spatial units. The SCI is calculated as \citep{bailey2018social}:

\begin{equation} \label{SCI}
        SC_{i,j} = \frac{FB\_Connections_{i,j}}{FB\_Users_{i} \times FB\_Users_{j}}
\end{equation}

where $FB\_Connections_{i,j}$ is the total number of Facebook friendship connections between users in location $i$ and users in location $j$. Similarly, $FB\_Users_{i}$ and $FB\_Users_{j}$ represent the total number of Facebook users in locations $i$ and $j$, respectively. Facebook determines a user's geographic location based on their profile and connection locations. Because the numerator counts mutual friendships and the denominator is the product of user counts in the two areas, SCI is mathematically symmetric, i.e.\ \(SC_{i,j}=SC_{j,i}\). The final index is scaled to a range between 1 to 1,000,000,000. The state-level dataset covers 50 states and Washington, D.C. in the U.S., which means each state has 50 values of SCIs representing the strength of connections to other states. It is worth reiterating that these connections include, but are not limited to, geographic proximity. The dataset is available for other countries as well, providing a valuable source for similar modeling in other regions of the world, including data-poor regions. In its current release, SCI provides connectedness indices for thousands of sub-national units across more than 200 jurisdictions worldwide, a spatial coverage unmatched by commercial cell-phone mobility products \citep{ilin2021public, bailey2018social}. 

\citet{kuchler2022jue} introduced a metric termed Social Proximity to Cases (SPC) to quantify the impact of social connections on disease spread. This metric captures the evolving spatial characteristics of the epidemic through a weighted sum of cases in connected counties, with weights determined by the SCI. In pursuit of our goal of forecasting incident hospitalizations, we extend upon SPC to define Social Proximity to Hospitalizations (SPH). SPH represents a weighted sum of hospitalizations in connected states, encapsulating the spatial spill-over effect, i.e., the transmission of infectious disease and resulting hospitalizations from connected states. Specifically, the SPH is calculated as:

\begin{equation} \label{SPH_equation}
    SPH_{i,t} = \sum_{j \in C} Hospitalization\_rate_{j,t} \times \frac{SC_{i,j}}{\sum_{h \in C} SC_{i,h}}
\end{equation}

where $Hospitalization\_rate_{j,t}$ represents the number of hospitalizations caused by COVID-19 per 10,000 people in state $j$ at time $t$. $SC_{i,j}$ denotes the SCI between state $i$ and $j$. $C$ is the set of all states connected to state $i$. In Eq.~\eqref{SPH_equation}, we first row-normalized the SCI for each focal state $i$. This normalization converts the symmetric SCI matrix into a directional weight vector that represents the proportion of social connections from state $i$ to every other state $j$. Consequently, SPH can be interpreted as the weighted sum of hospitalization rates in other states, weighted by how socially connected those states are to the target state $i$. While SCI remains a static annual index, SPH captures pandemic dynamics by integrating daily updated hospitalization data at time $t$. Time-series SPH values are computed for each state, alongside incident cases and hospitalizations, and are updated at each time step to effectively capture both temporal and spatial changes.

\begin{figure}[h]
\centering
\vspace{-6pt}
\includegraphics[width=14cm]{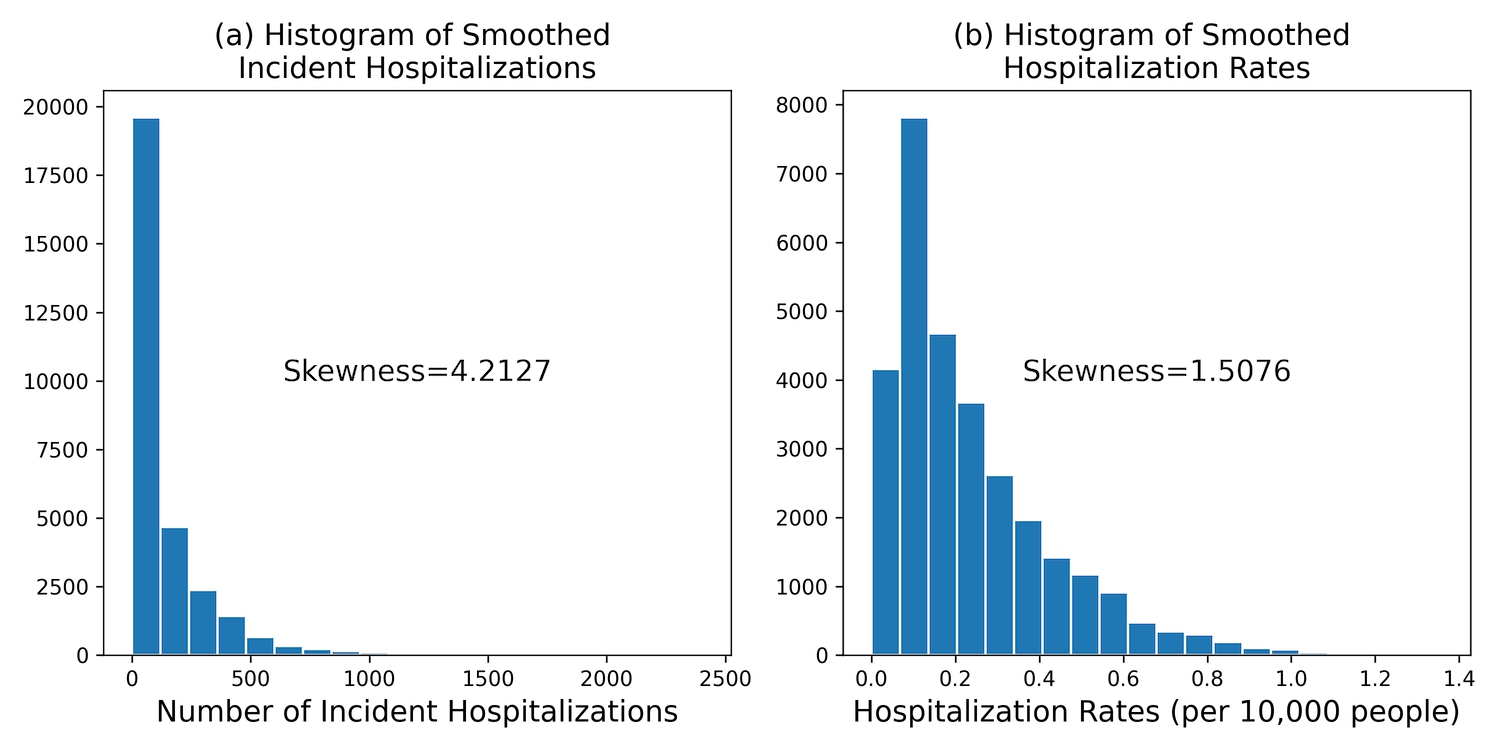}
\caption{Histogram and Skewness Index of Raw Incident Hospitalizations and Hospitalization Rates by March 12, 2022, for all 51 states (around our evaluation period of the Omicron surge). Rates are less skewed, allowing for a better capture of spatial spillovers, independent of population size. } \label{raw_rate}
\vspace{-6pt}
\end{figure}

Our approach differed from most others in that we first converted raw numbers into rates (incident cases or hospitalizations divided by state population), then rescaled those rates to the [-1,1] range using a MinMax scaler before feeding them into the model. Compared to raw numbers, rates are less skewed (see Fig.\ref{raw_rate}), enabling the model to learn more uniformly from all states of varying populations instead of focusing on states with larger populations. Our exploratory experiments showed better predictive performance using rates over raw incidence, and more importantly, better leveraging of the SPH features regardless of state population size. 

\subsection{Model Architecture and Hyperparameters}

\begin{figure}[h]
\centering
\vspace{-6pt}
\includegraphics[width=14cm]{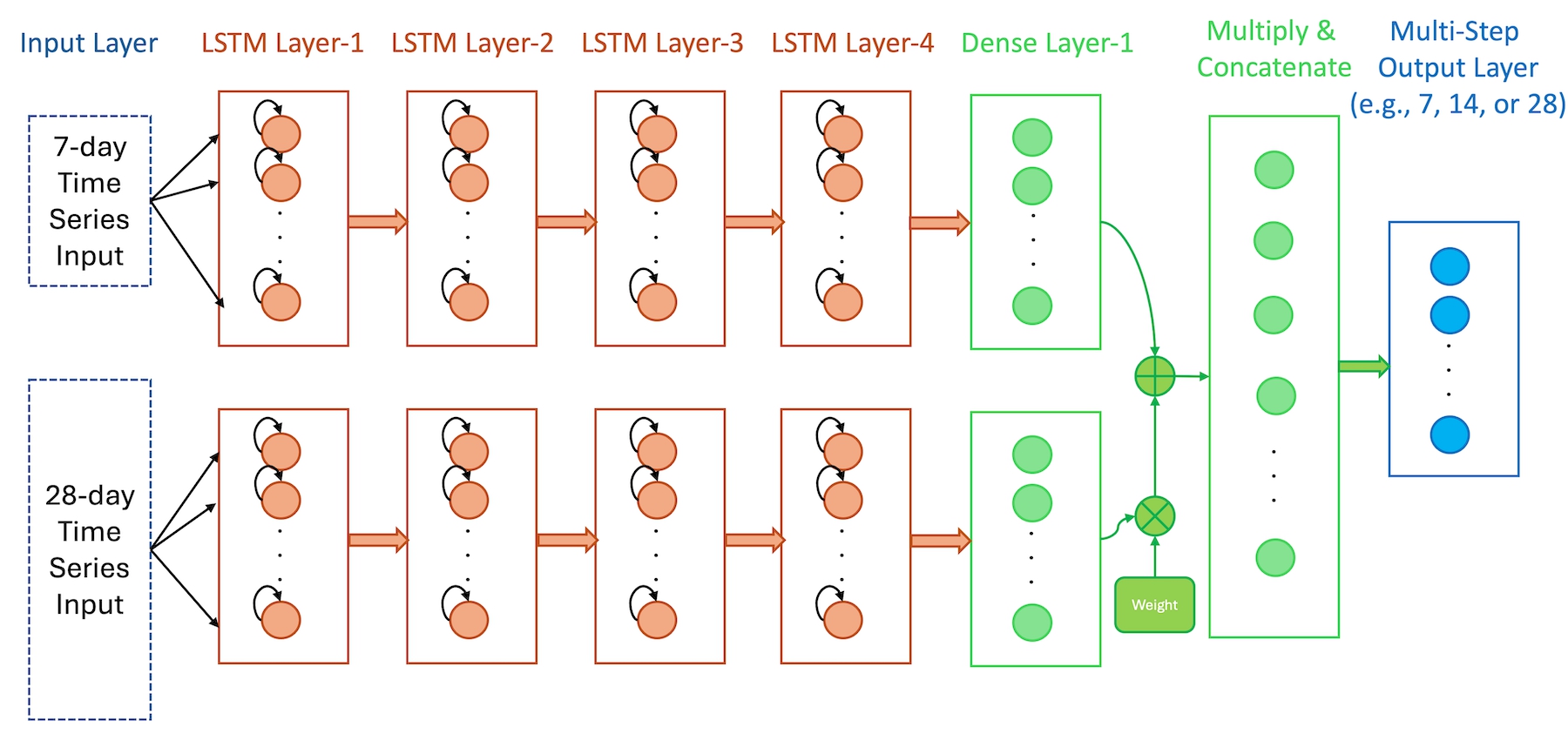}
\caption{Architecture of SLSTM, i.e., one of the parallel network members of our multi-horizon ensemble. The learnable parameter (weight) denoted in the green box balances information learned from the 7- and 28-day input branches and is learned for each ensemble member and forecast date separately.} \label{SLSTM_arch}
\vspace{-6pt}
\end{figure}

We developed a parallel stacked Long Short-Term Memory (LSTM) network with spatial features to forecast multi-horizon COVID-19 hospitalizations, which we refer to as SLSTM. Learning multi-scale temporal dependencies is crucial for effectively modeling the dynamics of the pandemic, which can vary across different phases and variants. The parallel-stream design captures both short-term (7-day) trends and long-term (28-day) patterns, allowing the model to balance these temporal scales and adapt to the dynamic nature of hospitalization trends. Preliminary tests with various subperiod combinations (e.g., 7 \& 14 days, 7 \& 21 days, or 14 \& 28 days) revealed that pairing 7 and 28 days yielded the best performance. The parallel-stream design incorporates two architecturally-identical stacked LSTM networks, with each network containing five hidden layers, as depicted in Fig.\ref{SLSTM_arch}. These networks comprise four LSTM layers with decreasing neuron counts (256, 128, 128, and 128) and a final dense layer with 64 neurons. These networks are designed to capture both short-term (7-day) trends and long-term (28-day) patterns by processing 3-channel multivariate time series arrays with different temporal lag lengths. This design allows the model to balance temporal scales and adapt to dynamic hospitalization trends by fusing short-term and long-term information. To balance short- and long-term temporal dependencies, we introduced a learnable weight, $W$, which was initialized to 1.0 and optimized via backpropagation. This weight modulates the short-term embedding before concatenation with the long-term embedding: 
\begin{align*}
    h_{\text{fused}}=\text{Concat} (W\cdot h_{\text{short}}, h_{\text{long}})
\end{align*}
This design allows the model to learn the optimal emphasis on recent versus extended temporal trends in the inputs. The fused vector $h_{\text{fused}}$ is then passed through a final dense layer to produce the final predictions. The output layer dimension of each network is determined by the ensemble strategy and forecast horizon lengths, which are 7, 14, and 28 days as discussed in Section \ref{ensemble_strategy}. A detailed pseudocode summary of the model architecture has been added in Appendix \ref{model_pseudocode}.

 We added early stopping to avoid overfitting. The model utilizes Quantile Loss as the loss function (described in Section \ref{quantile_loss}) and is optimized by Adam optimizer \citep{kingma2014adam} with an initial learning rate of 0.0008, trained with a batch size of 64. 

\subsection{Quantile Loss} \label{quantile_loss}

To quantify the uncertainty in our forecast and in accordance with the Hub's operational and submission requirements, our SLSTM model predicts 23 quantile values for each time-step, making it a quantile regression \citep{wen2017multi, wei2006quantile, koenker1978regression, kocherginsky2005practical}. In quantile regression, models are trained to minimize the total Quantile Loss (QL), rather than a single loss. The QL for an individual data point is defined as:

\begin{equation} \label{ql_loss}
    L_{q}(y, \hat{y}) = q(y-\hat{y})_{+} + (1-q)(\hat{y}-y)_{+}
\end{equation}

where $(\cdot)_{+} = \max(0, \cdot)$. Here, $q$ denotes the quantile level, ranging from 0 to 1. Theoretically, when $q=0.5$, the QL simply equals the Mean Absolute Error (MAE) for the median of the prediction interval. For multi-step and multi-quantile outputs, the total loss being minimized is:

\begin{equation}
    L_{Total} = \frac{1}{q \cdot k}\sum_{q}\sum_{k}L_{q}(y_{t+k}, \hat{y}^{(q)}_{t+k})
\end{equation}

where $k$ is the number of forecast time-steps, $q$ is the number of quantiles, and $t$ is the forecast date. As required by the Hub, $k=28(\text{days})$, and $q=23$, with values as follows: 0.010, 0.025, 0.050, 0.100, 0.150, 0.200, 0.250, 0.300, 0.350, 0.400, 0.450, 0.500, 0.550, 0.600, 0.650, 0.700, 0.750, 0.800, 0.850, 0.900, 0.950, 0.975, and 0.990.

\subsection{Multi-Horizon Ensembling Strategy} \label{ensemble_strategy}

There are four main strategies for multi-step time series forecasting: direct multi-step, recursive multi-step, direct-recursive hybrid, and multiple output strategy. The direct multi-step strategy trains a separate model for each forecast time step, which is computationally burdensome and tedious to maintain. The recursive multi-step strategy only trains one model and uses its predictions as inputs to predict subsequent time steps. This is less computationally demanding than the direct strategy; however, this approach propagates errors along the time steps \citep{marcellino2006comparison}. In the direct-recursive hybrid strategy, a separate model is trained for each time step to be predicted, but each model may also use the predictions from models predicting prior time steps. The final strategy, the multiple output strategy, predicts the entire forecast sequence (i.e., all output time steps) at once. Multiple-output models are more complex, which means they are harder to converge; however, their forecasts are more consistent across time steps and have smaller total errors. As the multi-output models are optimized over the entire sequence, the larger time-span predictions in the sequence may have larger errors than other strategies, where individual time-step errors are minimized separately, but more consistent across the entire time-span.  Lastly, all deep learning models are trained stochastically (using variations of stochastic gradient descent), resulting in varying predictions for models trained with different initial random seeds. 

\begin{figure}[h]
\centering
\vspace{-6pt}
\includegraphics[width=14cm]{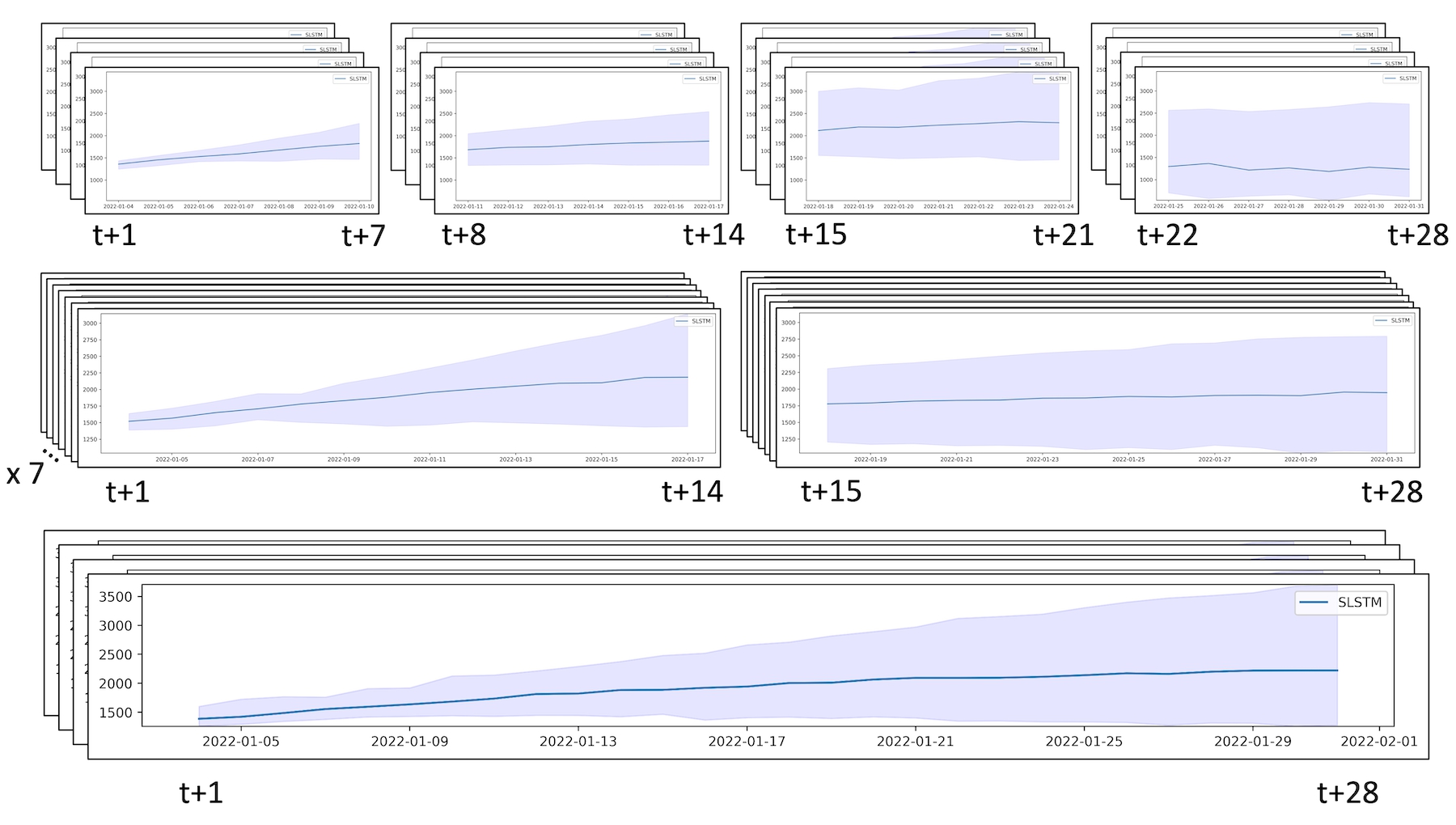}
\caption{Ensembling Strategy. For each time-step $t$ in the output, there are 15 predictions for each output quantile, using 7-day, 14-day, and 28-day output sequences to balance between consistency and error optimization per sequence.} \label{ensemble_strategy_fig}
\vspace{-6pt}
\end{figure}

Our SLSTM uses a multiple-output strategy. To counter the disadvantages of the multi-output strategy, i.e., to reduce the error and variability of stochastic predictions while maintaining the consistency of different output sequences, we devised a multi-horizon ensemble strategy as follows. We designed the ensemble such that for each 28 time-step in the forecast sequence, 15 different models generate a forecast for each quantile and point prediction (see Fig. \ref{ensemble_strategy_fig}). 

Among the 15 prediction members for each time step, 4 models have a 7-day sequence forecast horizon, with each of those 4 having a different random initial seed. Referring to Fig. \ref{ensemble_strategy_fig}, it is worth noting that we staggered four subgroups of such 7-day forecast output models such that the first subgroup forecasts hospitalizations of $t+1, \dots, t+7$; the second subgroup forecasts the next 7 days of $t+8, \dots, t+14$; the third subgroup forecasts the next 7 days of $t+15, \dots, t+21$; the fourth subgroup forecast the last 7 days: $t+22, \dots, t+28$. These 7-day forecast models are expected to have less error over the 7-day output sequence (compared to optimizing a model on 28 days); however, the forecasts made by subsequent subgroups may suffer inconsistency, e.g., the difference in predictions between $t+7$ from the first subgroup and $t+8$ from the second subgroup might be large. 

Therefore, we also included models with larger output sequences in the ensemble. Similarly, two staggered subgroups with seven members each are trained to forecast 14-day sequences, with seven members predicting $t+1, \dots, t+14$, and seven members predicting $t+15, \dots, t+28$. 

Lastly, we also included four members with a direct multi-output 28-day forecast sequence. This approach enables us to generate forecasts for the entire 28-day period in a single step, thereby maintaining consistency.

Once all forecasts are generated, the final ensemble output is determined by calculating the median of all 15 members for each point/quantile prediction to ignore outliers. This ensures that the ensemble output remains representative of the collective predictions while minimizing the influence of potential outliers. Furthermore, if necessary, quantile values are reordered to maintain consistency and accuracy across the forecast sequence.

\subsection{Model Training}

We employed a walk-forward validation approach to train the model, utilizing all available data up to and including the forecast date, which is made accessible before the Hub submission deadline every week. Models for each forecast date were trained using the most recent 15 months of data, capturing the latest transmission patterns, hospitalization dynamics, and population immunity. We used spatial cross-validation for training our models, and the final testing was conducted purely on forecasting dates during periods that occurred after the training set period. We configured our training strategy such that for each forecast date, it selected three states exhibiting the lowest, median, and highest hospitalization rates (a criterion that changes weekly, so different states occupy each role over time) to serve as our validation for early stopping. This spatial validation strategy ensures that the model is trained while consistently validated against a diverse range of epidemic dynamics- specifically, states at increasing, decreasing, and persisting stages of epidemic spread \citep{lever2016points}. This ensures that the model is not overfitting to increasing or decreasing trends. The remaining states, with varying hospitalization rate trends are used for training the model. This method allows us to implement early stopping to prevent overfitting and test the model's generalizability across a representative cross-section of epidemic conditions. After training is completed, the model is tested on time periods following the training period (with no overlap), to ensure the model’s ability to forecast the future. 

During training, the time-series SPH values, along with incident cases and hospitalization rates, were first normalized to the range [-1, 1] using the MinMax Scaling technique. These normalized features were then sequentially fed into the LSTM model, which processes the three input features over successive time steps. This integration enhances the model's ability to predict hospitalizations based not only on historical patterns within the same state, but also on historical patterns across connected states, with the strength of connectedness measured by social connectivity.

\begin{figure}[h]
\centering
\vspace{-6pt}
\includegraphics[width=12cm]{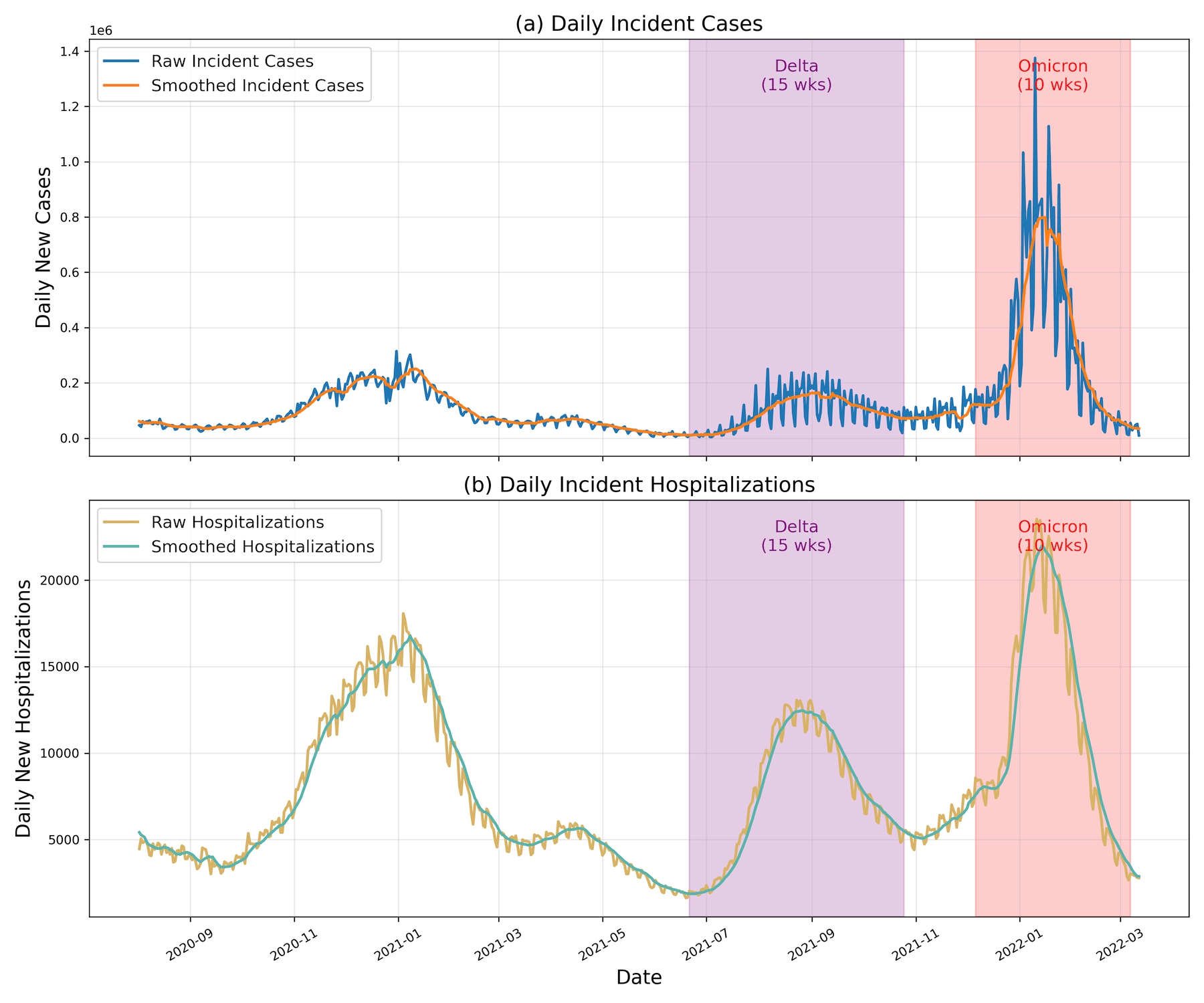}
\caption{Raw and smoothed daily U.S. COVID-19 cases (top) and hospitalizations (bottom). Purple and red bands mark the Delta (15 weeks, June 21 - September 27, 2021) and Omicron (10 weeks, December 6 - February 7, 2022) evaluation windows; each band extends 28 days beyond its final forecast date to show the validation horizon.} \label{cases_hosp}
\vspace{-6pt}
\end{figure}

The data exhibit inconsistencies stemming from several factors: differences in reporting frequency between states (daily versus weekly release of daily admissions), changing frequency (from daily to weekly), reporting errors (such as non-positive values), dips caused by under-reporting during weekends and holidays, followed by over-reporting spikes, and retroactive distribution of incidence due to glitches in initial reports. To mitigate these issues, we applied a 7-day rolling average to both cases and hospitalization data (shown in Fig. \ref{cases_hosp}). Recognizing that forecasts may be adversely impacted by data irregularities and anomalies \citep{lucas2023spatiotemporal}, we selected smoothed hospitalizations as the evaluation target instead of raw values, believing that this 7-day average better reflects reality than daily reports, which are frequently influenced by weekend and holiday reporting lags.

\subsection{Model Evaluation}
While SLSTM was employed during various phases of the pandemic to submit forecasts to the Hub, for the results presented in this paper, we utilized data from July 27, 2020, to March 7, 2022, a period that includes the two most severe U.S. surges: the Delta and Omicron waves. The Delta evaluation period spans 15 weeks (Jun 21 - Sep 27, 2021), capturing the rapid rise from late July to early August, the peak from late August to early September, and the decline throughout September. The second evaluation period spans 10 weeks, primarily focusing on the Omicron wave in the U.S., encompassing the substantial surge in cases in late December 2021, the peak in mid-January 2022, and the subsequent decline in late January 2022. These periods were chosen for model evaluation to test generalizability, as it covered the latest major wave of cases and hospitalizations in the U.S., which challenged forecast models given the explosive spread of the Delta and Omicron variant \citep{lopez2024challenges} and the abrupt changes in hospitalization dynamics between the two waves. Although the Pearson correlation between cases and hospitalizations remained strong at 0.74 during the Omicron wave (December 6, 2021, to February 7, 2022)---even higher than during the Delta wave (0.62)---the case-to-hospitalization ratio nearly doubled (28.52 vs. 13.40). This underscores that, while overall trends in cases and hospitalizations can remain broadly aligned, the magnitude shifted significantly during Omicron, with far more infections per resulting hospitalization---a change likely driven by variant-specific factors and the rise of at-home testing. Consequently, forecasting models must focus on direct hospitalization indicators rather than solely rely on infection trends \citep{fuss2022difference}.

We evaluated and compared our model performance using three commonly used point-forecast metrics and one probabilistic evaluation score.

Mean Absolute Error (MAE) at time $t$ is calculated as:

\vspace{-10pt}
\begin{equation}
    \text{MAE}_{t} = \frac{\sum_{i=1}^{N}|y_{i,t} - \hat{y}_{i,t}|}{N}
\end{equation}
\vspace{-10pt}

where $N$ is the total number of forecasts made at time $t$ (in our cases, 51 for U.S. States and D.C.), and $y_{i,t}$ and $\hat{y}_{i,t}$ are the true value and the predicted value, respectively, for prediction $i$ at time-step (day) $t$, which varies between the 1st day to 28th day after the Forecast Day, in accordance with the hub.

Mean Absolute Percentage Error (MAPE) measures accuracy as a percentage and is scale-independent and easy to interpret. However, MAPE may produce infinite and undefined values for zero or close-to-zero true values. As our target is state-level values, it is highly unlikely that there will be zero or close-to-zero values. Formally, MAPE at time $t$ is defined as:

\vspace{-10pt}
\begin{equation}
    \text{MAPE}_{t} = \frac{100}{N}\sum_{i=1}^{N}\frac{|y_{i,t} - \hat{y}_{i,t}|}{y_{i,t}}
\end{equation}
\vspace{-10pt}

where the variables have the same interpretations as above.

Root Mean Squared Error (RMSE) is more sensitive to large errors due to the squaring operation and is calculated as:
\vspace{-10pt}
\begin{equation}
    \text{RMSE}_{t} = \sqrt{\frac{\sum_{i=1}^{N}(y_{i,t}-\hat{y}_{i,t})^{2}}{N}}
\end{equation}
\vspace{-10pt}

In addition to point-based errors, we evaluated the Weighted Interval Score (WIS), a \textit{proper scoring rule} designed to evaluate probabilistic forecasts that approximates the continuous ranked probability score. This score was adopted by the Hub as the main evaluation metric \citep{cramer2022evaluation}. WIS rewards forecasts that are both sharp (narrow intervals) and well-calibrated (intervals that cover the truth  at multiple quantile levels). Let $m$ be the predictive median and $\{(l_k,u_k)\}_{k=1}^{K}$ the $K$ central prediction intervals with nominal coverages $1-\alpha_k$. Then

\begin{equation}\label{eq:wis}
    \text{WIS}_{\alpha_{1:k}} (y) = \frac{1}{K+1/2}[\frac{1}{2}|y-m|+\sum_{k=1}^{K}w_{k}\text{IS}_{\alpha_{k}}(y)], w_{k}=\frac{\alpha_{k}}{2}
\end{equation}
where $\text{IS}_{\alpha_k}$ is the interval score of the $(1-\alpha_k)$ interval , i.e.\ the sum of its dispersion (width) and penalty terms for under- and over-prediction (full formula in Appendix~\ref{appdendix:wis}). Lower WIS values indicate tighter and more reliable predictive distributions.

\section{Results}

\subsection{Predictive Power of Spatial Features}

\begin{figure}[ht]
\centering
\includegraphics[width=12cm]{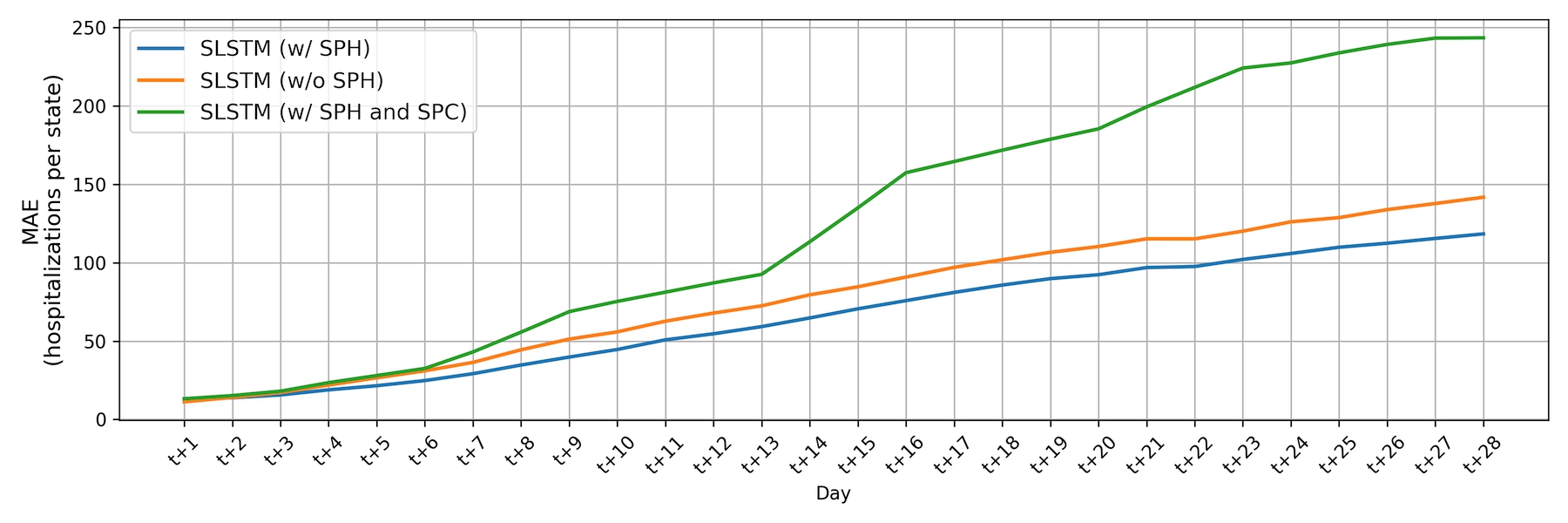}
\caption{Average MAE over the 10-week evaluation period for models with SPH, without SPH, and with both SPH and SPC spatial features.} \label{feature_comparison_mae}
\end{figure}

To evaluate the predictive performance of the Facebook-derived spatiotemporal feature (SPH), we trained three identical multi-horizon ensemble parallel LSTMs: one with all input features (smoothed cases, smoothed hospitalizations, and SPH), one excluding SPH, and a third model incorporating both SPH and SPC. The results in Table \ref{avg_sph_table}, Fig. \ref{feature_comparison_mae}, and Fig. \ref{feature_comparison_mape_rmse} demonstrate that the model with SPH has lower errors than the model without SPH across almost the entire 28-day horizon, except the first day in terms of MAE and MAPE, and the first two days in RMSE (the full daily comparison is available in Table \ref{spatial_eval_table}). On average, the spatial model outperformed the non-spatial model by 363 hospitalizations per state over the 28-day forecasting horizon, with the largest difference observed on the $28^{th}$ day, where the model with SPH was approximately 23 hospitalizations per state more accurate in MAE.

Interestingly, the model incorporating both SPH and SPC did not show consistent improvement and, in fact, suffered degradation in performance in most cases, particularly beyond the first four weeks of the evaluation period. During the initial weeks, the SPC-enhanced model demonstrated comparable or slightly better performance, especially until December 27, 2021. However, its performance declined significantly at the turning point of January 7, 2022, and in the subsequent weeks of the decreasing hospitalization trend. This result aligns with the hypothesis that the relationship between infections and hospitalizations shifted markedly during the Omicron wave. The lack of improvement when including SPC suggests that the spatial information embedded in recorded cases during this period did not contribute to the predictive performance of hospitalization models during such dynamic pandemic phases, especially in the presence of at-home testing.

\begin{table}[ht]
\tbl{Average errors of 28-day forecasts over the 10-week evaluation period for the models with SPH, without SPH, and with both SPH and SPC spatial features.}
{\begin{tabular}{lccc}\toprule
Model              & MAE   & MAPE  & RMSE   \\ \midrule
SLSTM (with SPH)    & 65.70 & 26.66 & 111.66 \\
SLSTM (without SPH) & 78.67 & 32.60 & 132.24 \\ 
SLSTM (with SPH and SPC) & 127.25 & 51.79 & 217.47\\ \bottomrule
\end{tabular}}
\label{avg_sph_table}
\end{table}

Beyond minimizing errors, we also found that the spatial models (with SPH) had a lower variance across the  15 ensemble members during the evaluation period. The variances of the spatial model were consistently lower than those of the non-spatial model, except for the final evaluation week (Table \ref{var_table}). The higher variances of the non-spatial model indicate that the model may be overfitting to the noise present in the data. On the contrary, the spatial model was able to capture the trends in data and has better generalizability.

\begin{table}[H]
\tbl{Models with spatial features exhibit lower variance.}
{\begin{tabular}{ccc}\toprule
Forecast Date & SLSTM (w/ SPH) Variance & LSTM (w/o SPH) Variance \\ \midrule
12/6/21       & 2284.33               & 129110.68              \\
12/13/21      & 5858.32               & 572274.18              \\
12/20/21      & 5620.26               & 163609.17              \\
12/27/21      & 27789.62              & 265972.34              \\
1/3/22        & 31333.79              & 163756.55              \\
1/10/22       & 34278.84              & 150484.13              \\
1/17/22       & 84840.44              & 941172.08              \\
1/24/22       & 246504.07             & 871446.52              \\
1/31/22       & 92367.33              & 400772.13              \\
2/7/22        & 36661.68              & 20888.87               \\ \bottomrule
\end{tabular}}
\label{var_table}
\end{table}

\subsection{Ablation Study of Multi-horizon Ensemble Strategy}

To assess the effectiveness of the multi-horizon ensemble strategy, we compared two different forecasting models: the SLSTM (Multi-horizon Ensemble) and the SLSTM (28-Day Direct). The multi-horizon ensemble model leveraged an ensemble strategy as described in Section \ref{ensemble_strategy}, where different members of the total 15 members learn on overlapping 7-day, 14-day, and 28-day periods. Conversely, the 28-day direct model consolidated the forecast into a single 28-day output span, with 15 models each predicting the entire 28-day period directly.  

\begin{figure}[H]
\centering
\includegraphics[width=10cm]{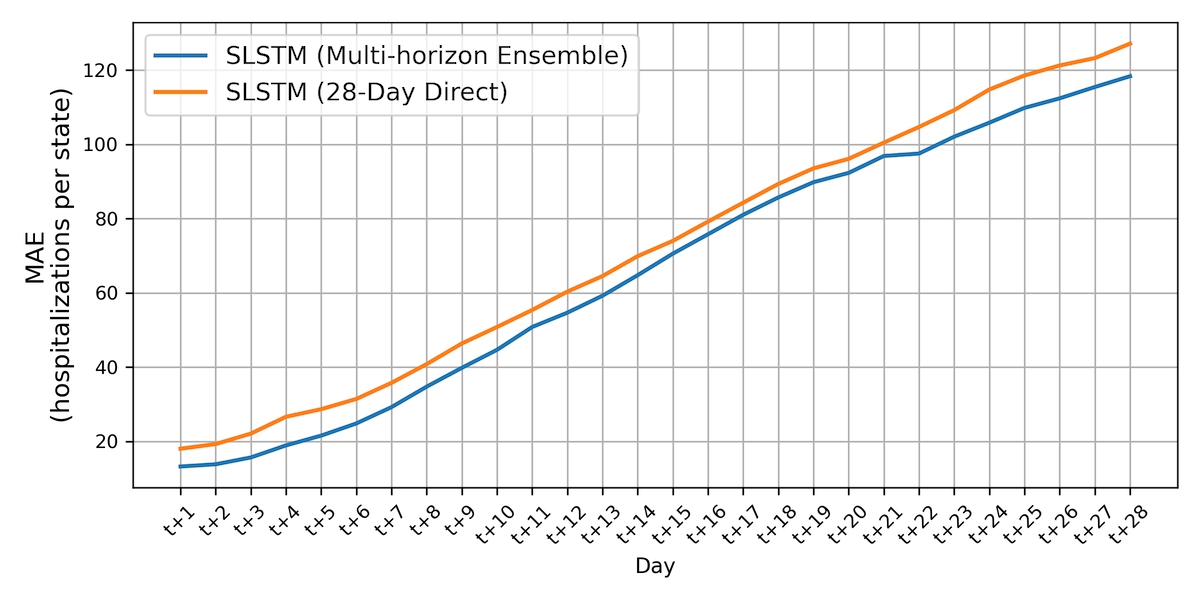}
\caption{Average MAE over the 10-week evaluation period for models using multi-horizon ensemble strategy and using 28-Day direct ensemble.} \label{multi-horizon_mae}
\end{figure}

The comparative performance of these models is visually presented in Fig. \ref{multi-horizon_mae}. The SLSTM (Multi-horizon Ensemble) consistently outperformed the SLSTM (28-Day Direct) across all time steps. Specifically, the multi-horizon strategy improved over the 28-day direct approach by an average of 167 hospitalizations per state throughout the evaluation period. This was also supported by lower Mean Absolute Percentage Error (MAPE) and Root Mean Squared Error (RMSE) values for the multi-horizon ensemble across the 28-day forecasting period in Fig. \ref{multi-horizon_mape_rmse}. 

The superior performance of the SLSTM (Multi-horizon Ensemble) can be attributed to its ability to reduce error over segments, as well as mitigate error propagation and drift inherent in longer-term direct forecasts. This strategy ensures a balance between errors in individual time steps and overall consistency for complex forecasting tasks such as predicting hospitalizations over an extended period.

\subsection{Comparison Against Baselines}

We compared the forecasts of SLSTM with three models from the COVID-19 Forecast Hub: \textit{COVIDhub-baseline}, \textit{COVIDhub-4\_week\_ensemble}, and \textit{COVIDhub-trained\_ensemble}. The first model is a persistence baseline, and the latter two are the top-performing ensemble models on the Hub (See Section \ref{ForecastHub})

\begin{figure}[ht]
\centering
\vspace{-6pt}
\includegraphics[width=14cm]{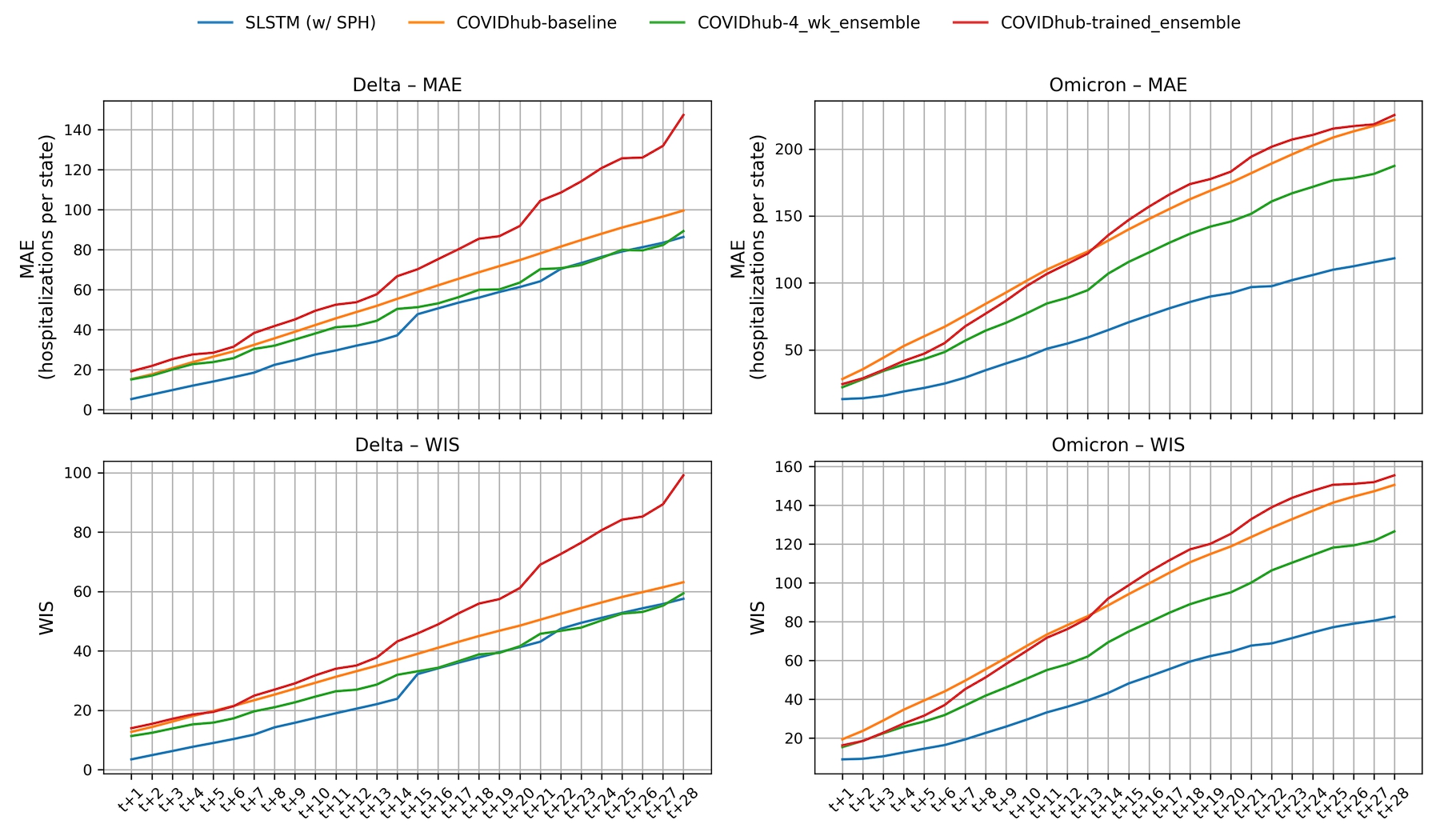}
\caption{Average prediction MAE and WIS over 28-day forecast horizon during the Delta (15 weekly forecasts, June 21–September 27, 2021) and Omicron (10 weekly forecasts, December 6, 2021–February 7, 2022) waves.} \label{model_comparison_main}
\vspace{-6pt}
\end{figure}

The results in Table \ref{tab:error_hub_comparison} and visualizations (in Fig. \ref{model_comparison_main} and Fig. \ref{model_comparison_appendix}) show that SLSTM consistently outperformed all other comparison models across the Delta and Omicron in all four evaluation metrics.  Daily comparisons (Table \ref{tbl:baseline_eval_combined_delta}) for the Delta wave reveal that SLSTM's advantage was larger for the first two-week horizon, while the performance in week 3 and 4 was only marginally better than that of the \textit{COVIDhub-4\_week\_ensemble}. This likely reflects the impact of limited training data for long-lead predictions during pandemic surges on deep learning-based models. During the Omicron wave, SLSTM not only preserved its short‑term advantage but also achieved substantial gains at longer horizons (weeks 3–4) (shown in Fig. \ref{tbl:baseline_eval_combined_omicron}). Specifically, SLSTM was, on average, 15, 30, 47, and 64 hospitalizations per state more accurate than the next best model, \textit{COVIDhub-4\_week\_ensemble}, for the $2^{nd}, 9^{th}, 16^{th}$, and $23^{rd}$ day, respectively. The \textit{COVIDhub-4\_week\_ensemble} had the second smallest error, and \textit{COVIDhub-trained\_ensemble} performed worse than the \textit{COVIDhub-baseline} in certain periods according to MAE and RMSE. This shows the value of deep learning-based models when sufficient, high-quality training data are available. 

\begin{table}[ht]
\tbl{Average 28-day forecast errors during the Delta
(15 weekly forecasts, June 21 – September 27, 2021) and Omicron (10 weekly forecasts, December 6, 2021– February 7, 2022) waves.
Lower values indicate better forecasts.}
{\scriptsize
\begin{tabular}{lcccccccc}
\toprule
& \multicolumn{4}{c}{\textbf{Delta wave}} 
& \multicolumn{4}{c}{\textbf{Omicron wave}} \\ 
\cmidrule(lr){2-5}\cmidrule(lr){6-9}
Model & MAE & MAPE & RMSE & WIS
      & MAE & MAPE & RMSE & WIS \\ \midrule
SLSTM (w/ SPH)             & 44.09 & 31.13 &  88.16 & 29.26
                           & 65.70 & 26.66 & 111.66 & 45.18 \\
\textit{COVIDhub-baseline} & 57.15  & 41.90 & 118.89 & 38.01
                           & 132.28 & 60.97 & 227.01 & 89.14 \\
\textit{COVIDhub-4\_week\_ensemble}& 50.13  & 34.15 & 105.96 & 32.96
                           & 108.08 & 41.27 & 193.42 & 71.26 \\
\textit{COVIDhub-trained\_ensemble}  & 72.45  & 43.08 & 151.56 & 48.12
                           & 133.34 & 47.42 & 242.96 & 90.91 \\ \bottomrule
\end{tabular}}
\label{tab:error_hub_comparison}
\end{table}

\begin{table}[ht]
\tbl{Decomposition of the 28‑day Weighted Interval Score (WIS)\textemdash%
average dispersion (Disp), penalties for under‑prediction (Under) and
over‑prediction (Over)\textemdash
for the Delta wave
(15 weekly forecasts, 21 Jun–27 Sep 2021) and the Omicron wave (10 weekly forecasts, December 6 2021 – February 7, 2022). Smaller values denote higher predictive performance.}
{\scriptsize
\begin{tabular}{lcccccc}
\toprule
            & \multicolumn{3}{c}{\textbf{Delta wave}} &
              \multicolumn{3}{c}{\textbf{Omicron wave}} \\ 
\cmidrule(lr){2-4}\cmidrule(lr){5-7}
\textbf{Model} & Disp & Under & Over & Disp & Under & Over \\ \midrule
SLSTM (w/ SPH)                &  6.88 & 13.63 &  8.75 & 10.51 & 24.42 & 10.26 \\ 
COVIDhub‑baseline    & 17.42 & 12.29 &  8.30 & 25.92 & 33.75 & 35.77 \\ 
COVIDhub‑4\_week\_ensemble  & 12.77 & 11.36 &  8.82 & 18.91 & 24.08 & 28.80 \\ 
COVIDhub‑trained\_ensemble  & 16.64 &  9.27 & 22.22 & 19.96 & 27.10 & 44.42 \\ 
\bottomrule
\end{tabular}}
\label{tab:wis_decomp_delta_omicron}
\end{table}

Looking at individual forecast dates within the evaluation period, SLSTM outperformed all other models in 7 of 10 forecast dates. The three dates that SLSTM had higher average prediction errors than \textit{COVIDhub-4\_week\_ensemble} occur during the increase (December 13, 2021, and December 20, 2021) and the decline (February 7, 2022) in hospitalizations. It is worth noting that SLSTM was substantially more accurate than the other three models around the peak of incident hospitalization during the Omicron wave (the middle of our evaluation period). This is due to comparison models overpredicting the peak, while SLSTM successfully forecasted the beginning of the decline of the Omicron wave. 

Beyond point forecasts, SLSTM also excelled in probabilistic forecasting. It had the lowest overall WIS in both waves (See Table~\ref{tab:wis_decomp_delta_omicron}). The improvement can reach 37\% over the next best model, \textit{COVIDhub-4\_week\_ensemble}, in the Omicron wave. When decomposing WIS into three components (shown in Table~\ref{tab:wis_decomp_delta_omicron}, \ref{tab:wis_decomposition_delta}, and \ref{tab:wis_decomposition_omicron}), SLSTM achieved the smallest dispersion in both waves, indicating tighter and more reliable prediction intervals. Consistent with the point‑based metrics, SLSTM led on short‑term horizons and maintained a slight advantage over the equally weighted ensemble (\textit{COVIDhub‑4\_week\_ensemble}) across the 28‑day window. During the Omicron wave, its under-prediction performance was similar to \textit{COVIDhub-4\_week\_ensemble}, but its over-prediction errors were significantly lower, showing fewer high-end misses and better performance in predicting turning point and decreasing phases. 

\begin{figure}[ht]
\centering
\vspace{-6pt}
\includegraphics[width=14cm]{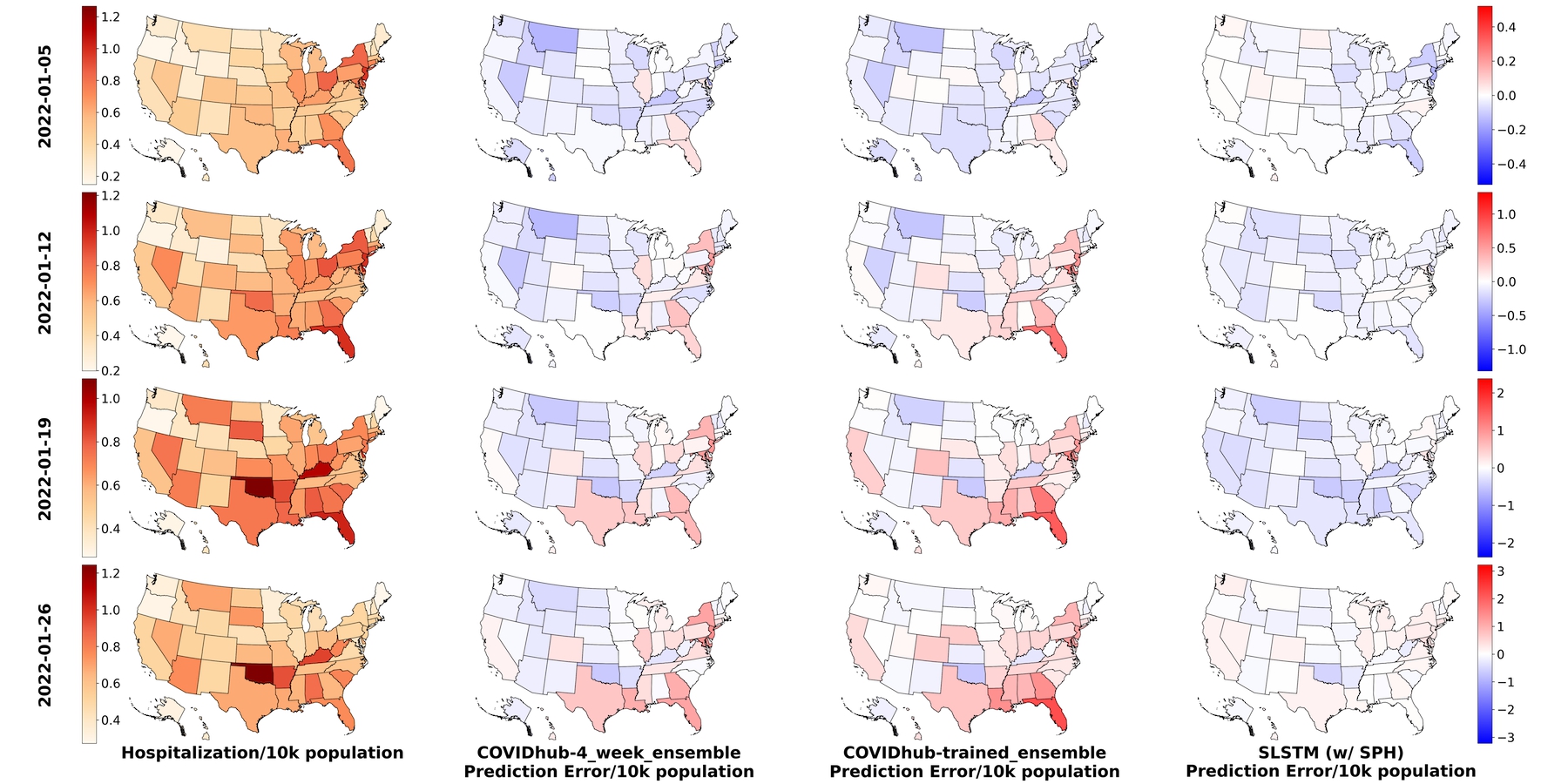}
\caption{Spatial distribution of hospitalization rates and normalized prediction errors of three baseline models and SLSTM on $2^{nd}, 9^{th}, 16^{th}$, and $23^{rd}$ days forecasted on January 3, 2022}\label{error_map}
\vspace{-6pt}
\end{figure}

Fig. \ref{error_map} presents the spatial distribution of the prediction errors on $2^{nd}, 9^{th}, 16^{th}$, and $23^{rd}$ days ahead as forecasted on January 3, 2022. All choropleth maps are normalized by population. Overall, SLSTM showed no systematic bias and had uniformly lower prediction errors across low-, median-, and high-incidence states than \textit{COVIDhub-4\_week\_ensemble} and \textit{COVIDhub-trained\_ensemble} on all four dates. The two ensemble models tended to overpredict in states with larger populations, while SLSTM had relatively smaller prediction errors. The patterns were most obvious in the forecasts for the last row ($23^{rd}$ day), with less spatial pattern left in the residuals.

\begin{figure}[ht]
\centering
\vspace{-6pt}
\includegraphics[width=14cm]{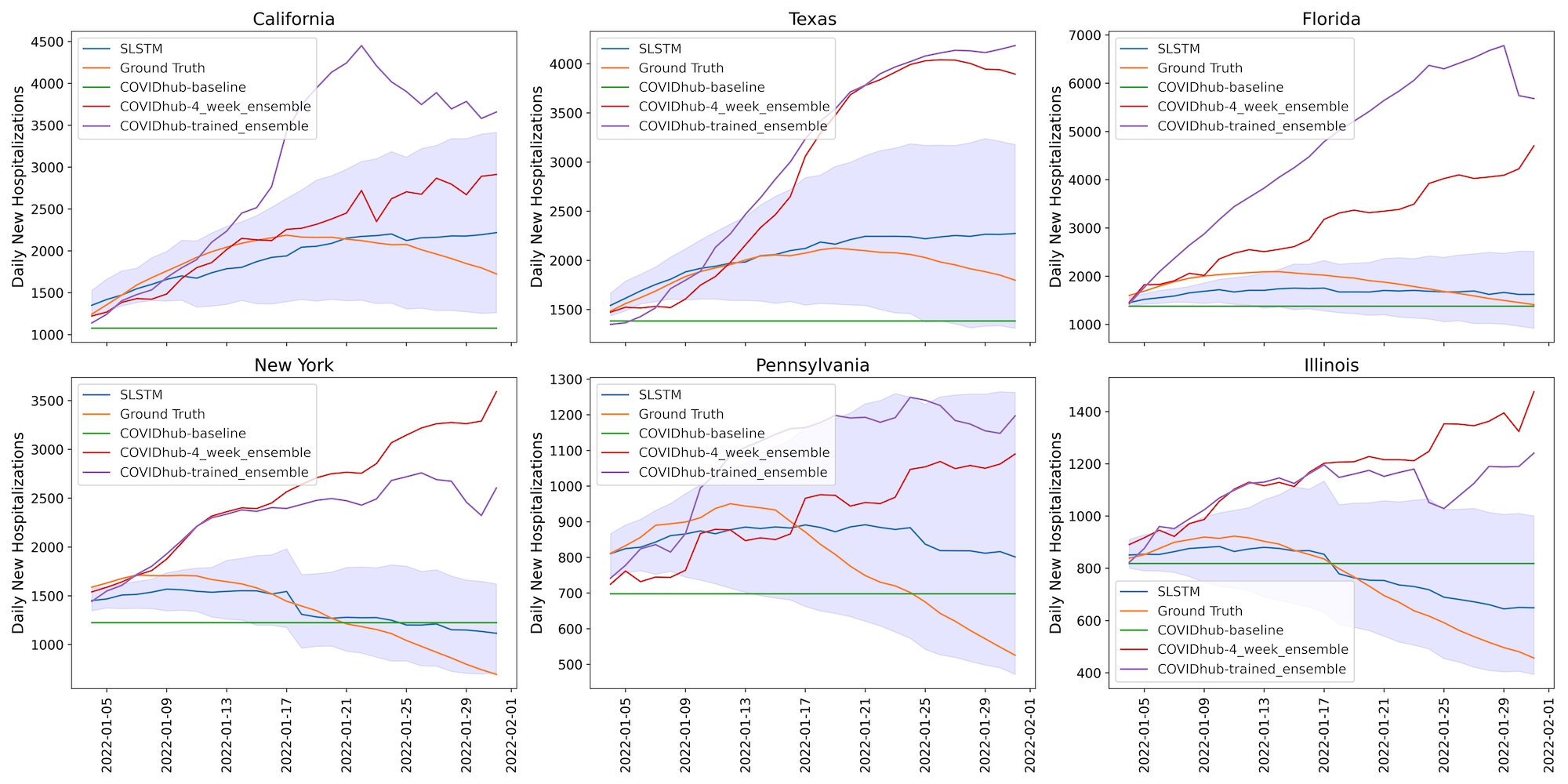}
\caption{COVID-19 hospitalization forecasts of SLSTM, \textit{COVIDhub-baseline}, \textit{COVIDhub-4\_week\_ensemble}, and \textit{COVIDhub-trained\_ensemble}. Blue shadow represents 95\% prediction intervals of SLSTM}\label{ts_interval_viz}
\vspace{-6pt}
\end{figure}

Fig. \ref{ts_interval_viz} compares hospitalization forecasts made on January 3, 2022, by SLSTM and the three comparison models. We chose the six states based on their population and hospitalization numbers. While it is evident from Fig. \ref{ts_interval_viz} that all models, including SLSTM, tended to overestimate hospitalization at the peak of the Omicron wave, our model was notably able to predict the non-linear changing trend of hospitalizations---transitioning from increasing to decreasing phases. The hospitalization curves of the six states at this time showed heterogeneity across the space, as eastern states were already declining, while California and Texas were about to reach their respective peaks.

\subsection{Prediction Intervals and Model Uncertainty}

The blue shaded band in Fig. \ref{ts_interval_viz} represents the 95\% prediction interval of SLSTM, generated using quantiles 0.025 and 0.975 in the model output for each output time-step. The uncertainty of forecasts increased further with the forecast horizon due to the inherent challenges of longer-term predictions. This increase in uncertainty is primarily driven by error propagation as the model increasingly relies on projections rather than direct observations. Additionally, changes in the characteristics of spatial units and variations in interventions over time introduced heterogeneity into the (historical) training and (future) testing sets, further contributing to uncertainty. While this is a perennial challenge in time-series forecasting, incorporating more dynamic features has the potential to help mitigate this effect. Future research should focus on developing more strategies to further reduce forecast uncertainty. Nevertheless, during the evaluation period and for all states, the 95\% prediction intervals of SLSTM forecasts included 90.20\% of ground truth values, while only 68.63\% and 70.24\% of ground truth values fell into the 95\% prediction intervals of \textit{COVIDhub-4\_week\_ensemble}, and \textit{COVIDhub-trained\_ensemble}, respectively. These qualitative findings align with the quantitative WIS results reported in Table \ref{tab:wis_decomp_delta_omicron}: SLSTM achieved the smallest dispersion and over-prediction penalties, confirming that its prediction intervals are tighter and more reliable. Together, these metrics demonstrated SLSTM's superior ability to quantify the prediction uncertainty and capture true values in the prediction intervals. 

\section{Discussion}

Our evaluation results highlighted the novel approach of incorporating spatiotemporal features derived from Facebook’s Social Connectedness Index into a multi-horizon, parallel LSTM framework. This unique integration significantly improved predictive performance and reduced variance and uncertainty in the hospitalization projection, particularly over longer-term horizons. 

Human mobility and the resulting interactions contribute to the transmission dynamics of infectious diseases \citep{changruenngam2020individual}. From a geospatial modeling point of view, this is similar to the spatial spillover effect in spatial lag modeling, where the target variable also influences the target variable in connected spatial units. However, accessing human mobility data can be expensive and limited due to privacy concerns and the high cost of data, making it simply impossible for many places and countries. At the time of this writing, many mobility datasets that were temporarily made available during the pandemic are still unavailable, even in the US post-pandemic. In contrast, Facebook's SCI is released and updated annually, remains freely accessible, and covers county-, state-, and country-level in more than 200 countries and territories, providing a stable, no-cost proxy for human interaction that persists beyond the pandemic period \citep{bailey2017measuring}. Beyond infectious disease modeling, SCI has been widely applied in studies of COVID-19 vaccine uptake \citep{basu2025social}, residential segregation \citep{nilforoshan2023human}, international migration \citep{minora2022migration}, and eco-innovation partnerships \citep{basak2024collaborating}, highlighting its broad applicability across public health, socio-economic, and behavioral domains. 

Human mobility from one spatial unit to another is highly correlated with social networks, and both follow spatial constraints, meaning that the majority of our trips or communications occur locally, with occasional longer journeys or calls to friends residing at a distance \citep{deville2016scaling}. Though the Social Connectedness Index (SCI) is a relatively static measurement based on social media friendships among pairs of spatial units, it is a good alternative proxy for mobility data such as those distributed by SafeGraph \citep{ilin2021public}---an advantage that is critical for low- and middle-income regions where commercial mobility data are either cost-prohibitive or entirely absent. In previous research, SCI-derived features demonstrated slightly stronger predictive power in short-term forecasting than SafeGraph’s mobility data, albeit using tree-based ensemble models \citep{vahedi2021spatiotemporal}.

One of the advantages of statistical forecasting for infectious diseases, such as our proposed model, is that such approaches can capture spatial heterogeneity and patterns in the data, even when variables are not directly included as predictor features. For instance, although vaccination clearly mitigates Omicron severity, reliable state-level vaccination data were unavailable during our study window (July 2020 - March 2022). Many jurisdictions did not publish daily dose counts or demographic breakdowns, and the CDC's data aggregation pipeline documented frequent gaps and de-duplication challenges \citep{bradley2021unrepresentative, scharf2024monitoring}. Therefore, rather than injecting a noisy, sparsely reported vaccination variable, our data-driven approach did not include predictor variables such as vaccination rates, especially since the vaccine versions, as well as the length and extent of immunity, are highly variable. Instead, we relied on incident cases, hospitalizations, and a SCI-derived spatiotemporal hospitalization lag feature, SPH. Because higher vaccine coverage lowers the infection-to-hospitalization ratio, those declined hospitalization signals are encoded in the time series and then propagated to socially connected states through the SCI weights, allowing SPH to capture vaccine-driven heterogeneity implicitly \citep{alvarez2022spatial, aslim2024vaccination}. Emerging evidence shows that county‑level vaccination uptake is strongly shaped by friendship networks, implying that SCI already embeds vaccination dynamics implicitly \citep{basu2025social}. Similarly, quantifying non-pharmaceutical interventions (NPIs) such as quarantine compliance, social distancing, and changes in health care provision is difficult and, therefore, challenging to incorporate into models directly \citep{iezadi2021effectiveness}. Instead, the statistical model learns from the patterns in data, where NPI efficacy is reflected in the time-series inputs of cases and hospitalizations in spatial units. 

Compared to compartmental models (i.e., the SIR family), the statistical approaches of deep learning update complex models with many learnable parameters using non-linear activation functions by feeding training data and employing backpropagation, rather than specifying differential equations and calibrating limited parameters. This flexibility enables the neural network to internalize time-varying factors, such as emerging variants, shifts in mobility, and spatially clustered vaccine uptake, via SPH, without requiring explicit corresponding covariates. Due to the high variability of SARS-CoV-2 variants and heterogeneity across space and time, the key indicator, $R_0$, varies between variants \citep{du2022reproduction}, spatial units \citep{thomas2020spatial}, and age groups \citep{monod2021age}. This is another indication that compartmental models are not complex enough to capture the heterogeneity of the population as well as virus variants across space and time, compared to deep learning models. However, $R_{0}$ is crucial for tracking changes in disease transmissibility, enabling decision-makers to evaluate the effectiveness of interventions. Other parameters make it possible to simulate different scenarios for policy-making, such as the slow waning of immunity and absence of a new variant, and fast waning, and with the emergence of a new variant \citep{borchering2021modeling}.

While raw incidence numbers or rates (incidence per 10k population) are used in compartmental models, there is no conclusive agreement on which works best \citep{mwalili2020seir, cooper2020sir}. In statistical approaches such as our method, it is better practice to use rates because raw incidence observations, which are highly skewed, lead to excessively large variance in estimates and worse predictive performance towards extreme values \citep{ribeiro2020imbalanced}. In our experiments, we found that using rates instead of raw numbers has better performance, making the model less sensitive to the population size of spatial units, especially in those states with extremely small or large population sizes. Interestingly, this is most evident in the spatiotemporal feature (i.e., the SPH feature), which has stronger predictive performance when rates are used, resulting in less model variance. 

To improve our model's performance, we also tested an additional feature based on Google Search Trends as an early signal of hospitalizations. We experimented with two groups of search trends, one including COVID-related terms (e.g., ``COVID-19", ``Coronavirus", and ``COVID Vaccine") and one narrowed down to COVID-related symptoms (e.g., ``loss of taste", ``loss of smell", ``cough", and ``fever"). Though both COVID-related terms and symptoms showed high correlations with the cases and hospitalizations, we found that incorporating Google search trends in our models only improved performance marginally during the increase period and deteriorated performance over the peak and decrease periods. We also found that COVID-related symptom search was more helpful than COVID-related terms during the increasing phase. One of the potential reasons that COVID-related symptoms did not improve over the peak and decreasing period may be the overlap with the flu season. COVID-19 and flu share similar symptoms, such as fever, cough, runny nose, and other symptoms, which may not correlate highly with COVID-19 cases and hospitalizations, such as ageusia, anosmia, and pneumonia. Given that the performance gain was marginal and only observed for a subset of our evaluation weeks, we decided not to incorporate this feature into the final models and results presented. Furthermore, the lack of an official API for accessing Google Search Trends, coupled with the inefficiency of using third-party tools such as pytrends for real-time forecasting due to their slow performance, reinforced our decision not to include this feature in the final models and subsequent results.

The improvements made by our model in predicting COVID-19 hospitalizations using SCI-derived features and a novel stacked architecture of LSTMs underscore its potential applicability to other infectious diseases with similar transmission dynamics and data collection practices. The inherent flexibility of our deep learning framework, which is capable of integrating diverse types of spatiotemporal data (e.g., county/state-level; daily/weekly), makes it well-suited for diseases like influenza that exhibit seasonal patterns and are influenced by human mobility and social behavior. The FluSight collaborative forecasting Hub, organized by the US CDC, was modeled after the COVID-19 Forecast Hub, and offers an ideal benchmark for testing the model's generalizability to influenza \citep{CDC2024, reich2019collaborative}. Future research will aim to evaluate the model's efficacy with data from other diseases.

We believe the techniques used in this study are not limited to the task-specific context of COVID-19 forecasting but can be generalized to other time-series forecasting problems. For example, the multi-horizon ensembling strategy and the parallel-stream LSTM model architecture are well-suited for applications such as air quality prediction, weather forecasting, and energy demand forecasting. Similarly, the quantile loss function is widely applicable to problems that require modeling uncertainty across different quantiles, such as traffic flow prediction.

\section{Conclusion}

In this paper, we presented a novel parallel Long Short-Term Memory (LSTM) framework with spatiotemporal features derived from Facebook's Social Connectedness Index to capture the spatial spillover effect. Our parallel architecture learns and combines short- and long-term temporal dependencies effectively. We also presented a novel multi-horizon ensembling strategy to balance between forecast consistency, uncertainty, and performance. We evaluated our forecasts on hospitalization reports during the Omicron surge and found that our approach outperforms multiple baselines over the 28-day horizon on average, with a high improvement margin towards the end. Long-term forecast, and the biggest average improvement on the $28^{th}$ day forecast can reach about 70 incident hospitalizations per state. Our evaluation results also demonstrated that our spatiotemporal feature, SPH, improved forecasts and reduced model variance. It is worth reiterating that the Omicron wave was the most challenging period for hospitalization forecasting models. In previous waves of the virus, hospitalizations typically followed case surges with a delay of approximately two to three weeks. With the change in coupling dynamics between cases and hospitalizations caused by the Omicron variant, which led to an explosive number of cases but fewer hospitalizations, the improvements achieved by our approach over baselines were notable. The strong predictive power of our approach is attributable to the novel spatiotemporal features integrated in LSTM, the use of incidence rates to decouple from population size, the parallel LSTM architecture with learnable balancing of short- and long-term dependencies, and a novel data-driven, multi-horizon ensembling strategy. 

The COVID-19 pandemic and the underperformance of forecasting models on the Forecast Hub \citep{lopez2024challenges} show the need for future research in this area. 
Despite the fact that statistical approaches have demonstrated high forecasting skills, fewer than one-third of the teams contributing to the Hub utilized statistical models. One potential reason is the legacy of compartmental models and their key parameters, such as $R_{0}$. Explainability and interoperability are valuable for epidemiologists to ensure the reliability of forecasting and to interpret the dynamics of epidemics. Therein lies the gap between the forecasting skill of statistical methods and the explainability of compartmental methods; future research should investigate ways to better combine the strengths of these two families of models. 

Another worthwhile direction lies in the use of spatiotemporal graph neural networks (GNNs). In our approach, the spatiotemporal feature that captures spatiotemporal spillover was formulated. In a GNN, nodes can represent spatial regions, while inter-region connectedness can be used as edge features \citep{kapoor2020examining}. GNNs have demonstrated their capability in node and graph classification tasks \citep{hamilton2017inductive}. However, their performance on regression tasks, such as those required by the Forecasting Hub, can be negatively affected due to oversmoothing in GNNs \citep{rusch2023survey}, indicating the need for further research. 

\section{Data and Codes Availability Statement}

The data and codes that support the findings of this study are available at https://github.com/geohai/covid-lstm-hosp.

\section*{Acknowledgement(s)}

We appreciate the detailed suggestions and comments from editors and anonymous reviewers.

\section*{Disclosure statement}

No potential conflict of interest is reported by the author(s).

\section*{Funding}

This work was supported by the Population Council, and the University of Colorado Population Center (CUPC) funded by Eunice Kennedy Shriver National Institute of Child Health \& Human Development of the National Institutes of Health (P2CHD066613). The content is solely the responsibility of the authors and does not reflect the views of the Population Council, or official views of the NIH, CUPC, or the University of Colorado.

\section*{Notes on contributor(s)}

\textbf{Zhongying Wang} is a graduate researcher in the Department of Geography at the University of Colorado Boulder, USA (E-mail: \href{mailto:zhongying.wang@colorado.edu}{zhongying.wang@colorado.edu}).  
His research focuses on geospatial data science and machine learning for public health.  
Wang led the data curation, software implementation, methodology design, formal analysis, visualization, and drafting of the manuscript. \\
\textbf{Thoai D. Ngo} is Professor in the Heilbrunn Department of Population and Family Health, Columbia University Mailman School of Public Health, USA (E-mail: \href{mailto:tn2571@cumc.columbia.edu}{tn2571@cumc.columbia.edu}). His work focuses on global public health and epidemiological research.  
Ngo conceived the study, secured funding, validated the analysis, and critically revised the manuscript. \\
\textbf{Benjamin Lucas} is a Post-doctoral Associate in the Department of Geography at the University of Colorado Boulder, USA (E-mail: \href{mailto:ben.lucas@nau.edu}{ben.lucas@nau.edu}). His interests include machine learning for remote-sensing applications. Lucas advised on the study methodology and contributed to the review and editing of the manuscript. \\
\textbf{Hamidreza Zoraghein}is a Research Scientist at the Population Council, New York, USA (E-mail: \href{mailto:hrz1365@gmail.com}{hrz1365@gmail.com}). He specialises in spatial data analysis and public-health applications. Zoraghein curated data resources and participated in the critical revision of the manuscript. \\
\textbf{Morteza Karimzadeh}: is Assistant Professor of Geography at the University of Colorado Boulder, USA (E-mail: \href{mailto:karimzadeh@colorado.edu}{karimzadeh@colorado.edu}). His research integrates geospatial data science, remote sensing and deep learning for public-health, social and environmental applications. Karimzadeh originated the study concept, directed the methodology, supervised the research team, oversaw project administration and funding acquisition, and contributed to both the original drafting and critical revision of the manuscript.



\bibliographystyle{tfv}
\bibliography{interacttfvsample}

\clearpage
 
\appendix

\section{SLSTM Model Architecture and Pseudocode} 
\label{model_pseudocode}

\begin{algorithm}
\caption{Short- and Long-Term LSTM Fusion for Quantile Prediction}
\begin{algorithmic}[1]
\Require $X_{\text{short}}$: short-term input sequence (e.g., 7 days)
\Require $X_{\text{long}}$: long-term input sequence (e.g., 28 days)
\Require $W$: learnable scalar weight
\Ensure $\hat{y}_q$: output predictions for each quantile $q$

\State $h_{\text{short}} \gets \text{LSTM\_stack\_short}(X_{\text{short}})$
\State $h_{\text{long}} \gets \text{LSTM\_stack\_long}(X_{\text{long}})$
\State $h_{\text{short\_proj}} \gets \text{DenseLayer}(h_{\text{short}})$
\State $h_{\text{long\_proj}} \gets \text{DenseLayer}(h_{\text{long}})$
\State $h_{\text{fused}} \gets \text{Concatenate}([W \cdot h_{\text{short\_proj}}, h_{\text{long\_proj}}])$
\For{each quantile $q$ in \texttt{quantiles}}
    \State $\hat{y}_q \gets \text{OutputLayer}(h_{\text{fused}})$
\EndFor
\end{algorithmic}
\end{algorithm}

\section{Weighted Interval Score (WIS) Details} \label{appdendix:wis}
\begin{enumerate}
    \item \textbf{Interval Score ($\text{IS}$)} for a single level\\
    For an observed value \(y\) and a central \((1-\alpha)\) prediction interval \(\bigl[l,u\bigr]\) \citep{gneiting2007strictly},
      \begin{equation}
      \text{IS}_{\alpha}(y)=
      \underbrace{(u-l)}_{\text{dispersion}} \;+\;
      \underbrace{\frac{2}{\alpha}\,(l-y)\,\mathbf 1_{\{y<l\}}}_{\text{underprediction}}
      \;+\;
      \underbrace{\frac{2}{\alpha}\,(y-u)\,\mathbf 1_{\{y>u\}}}_{\text{overprediction}}
      \label{eq:IS}
    \end{equation}
    where $\mathbf{1}$ is the indicator function, meaning that $\mathbf{1}(y<l) =1$ if $y<l$ and 0 otherwise. The terms $l$ and $u$ denote the $\alpha/2$ and $1-\alpha/2$ quantiles of predictions. 
    
    \item \textbf{Weighted aggregation across $K$ intervals and the median}\\
    With the predictive median $m$ and the same $K$ symmetric central
      intervals, the WIS is defined in Eq.~\eqref{eq:wis}.  We follow the Forecast Hub requirement of reporting 23 quantiles $\{0.01,0.025,0.05,0.10,\dots,0.95,0.975,0.99\}$.  Removing the median
      ($0.50$) leaves 22 quantiles, which form $K=11$ central intervals with $\alpha_k\in\{0.02,0.05,0.10,\dots,0.90\}$ used in that equation.
    \item \textbf{Implementation}\\
    All WIS values were computed with the \texttt{wis()} function from the \textbf{scoringutils} R package
      (v\,2.1.0) \citep{bosse2023scoring}. We kept the default settings \texttt{weigh = TRUE}, \texttt{count\_median\_twice = FALSE}, \texttt{separate\_results = FALSE}, and \texttt{na.rm = FALSE}. 
    
\end{enumerate}

\section{Comparative Analysis of Forecasting Results}

\begin{figure}[ht]
\centering
\includegraphics[width=12cm]{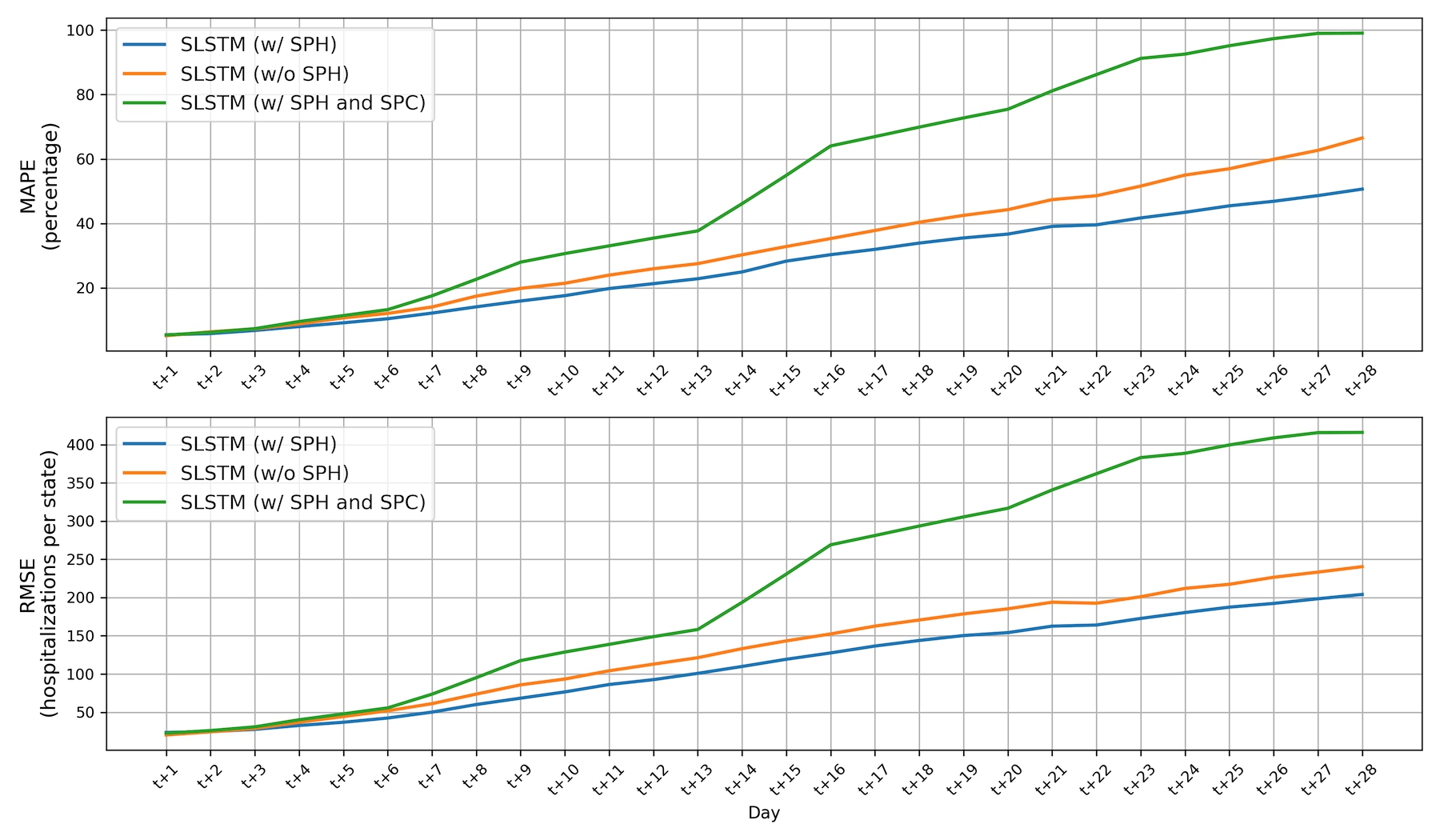}
\caption{Average MAPE, and RMSE over the 10-week Omicron evaluation period (December 6, 2021 – February 7, 2022) for models with SPH, without SPH, and with both SPH and SPC spatial features.} \label{feature_comparison_mape_rmse}
\end{figure}

\begin{figure}[H]
\centering
\includegraphics[width=12cm]{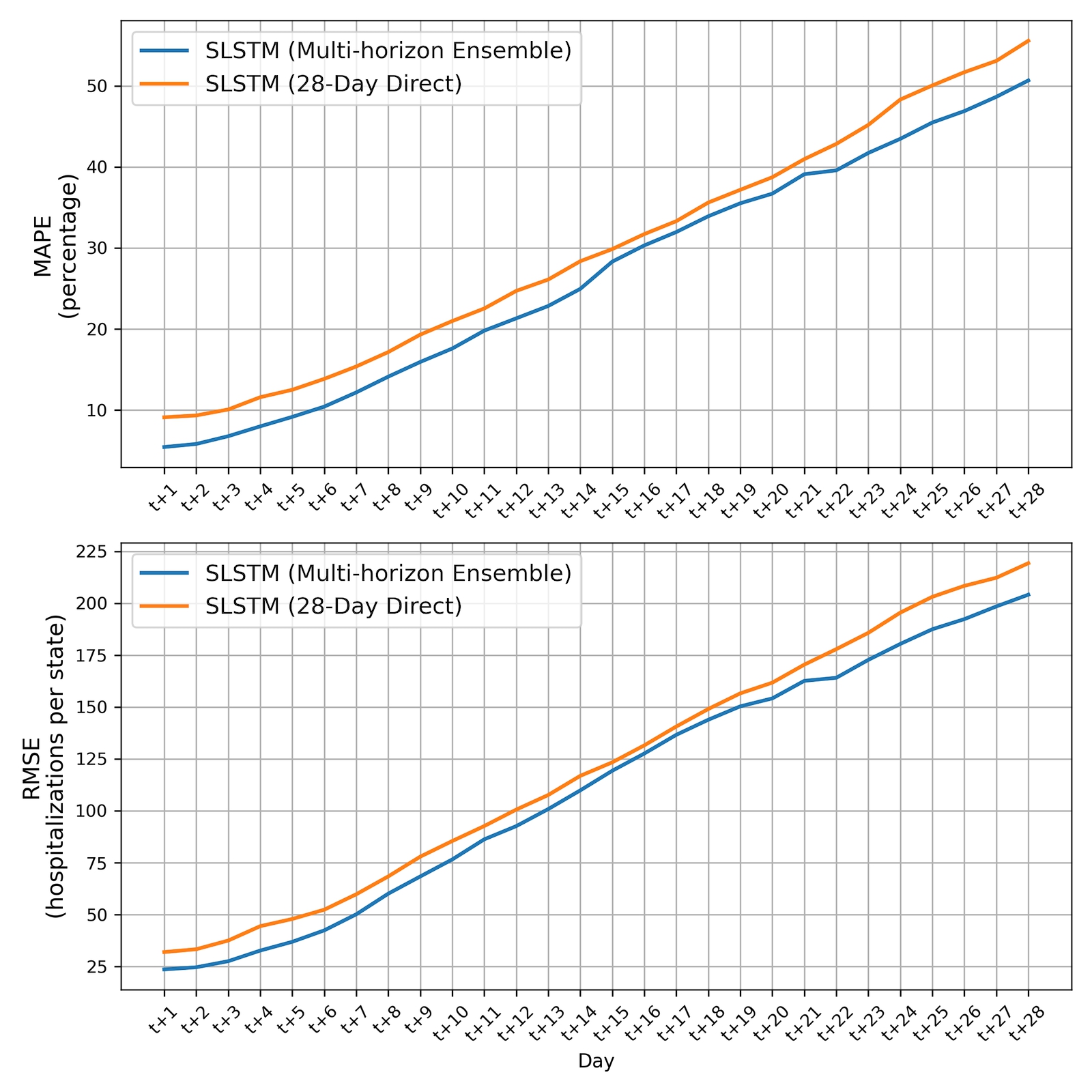}
\caption{Average MAPE, and RMSE over the 10-week Omicron evaluation period (December 6, 2021 – February 7, 2022) for models using multi-horizon ensemble strategy and using 28-Day direct ensemble.} \label{multi-horizon_mape_rmse}
\end{figure}

\begin{figure}[ht]
\centering
\vspace{-6pt}
\includegraphics[width=14cm]{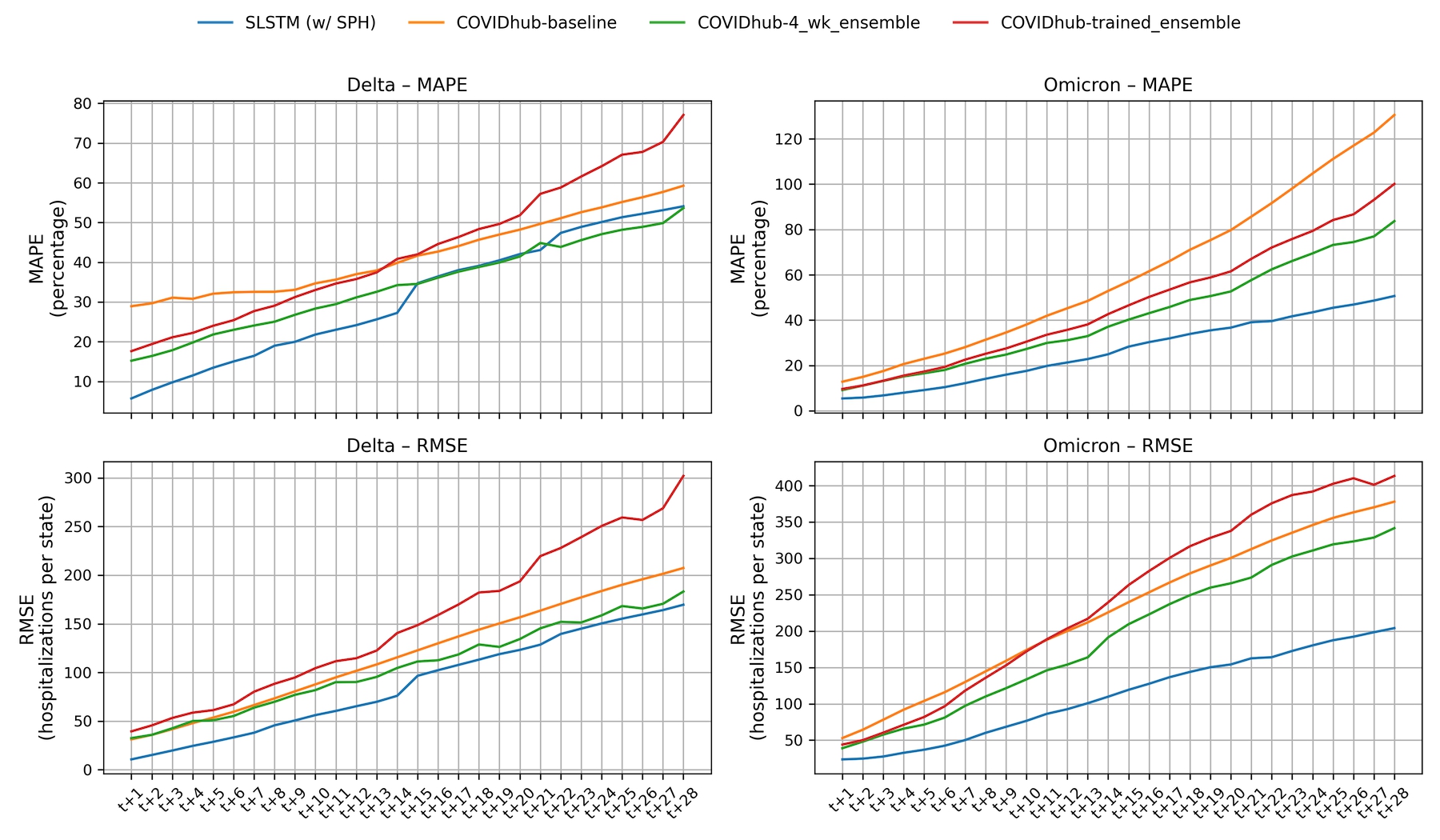}
\caption{Average prediction MAPE and RMSE over 28-day forecast horizon during the Delta (15 weekly forecasts, June 21 – September 27, 2021) and Omicron (10 weekly forecasts, December 6, 2021 – February 7, 2022) waves.} \label{model_comparison_appendix}
\vspace{-6pt}
\end{figure}

\begin{table}[H]
\tbl{Average MAE, MAPE, and RMSE of state-level predicted hospitalization incidence over 10 forecast dates (per week) in the Omicron evaluation period (December 6, 2021 – February 7, 2022).}
{\begin{tabular}{lcccccccccc}\toprule
    & \multicolumn{3}{c}{SLSTM (with SPH)}                                           & \multicolumn{3}{c}{SLSTM (without SPH)}                                       & \multicolumn{3}{c}{SLSTM (with SPH and SPC)} \\ \cmidrule{2-10} 
Day & \multicolumn{1}{l}{MAE} & \multicolumn{1}{l}{MAPE} & \multicolumn{1}{l}{RMSE} & \multicolumn{1}{l}{MAE} & \multicolumn{1}{l}{MAPE} & \multicolumn{1}{l}{RMSE} & \multicolumn{1}{l}{MAE} & \multicolumn{1}{l}{MAPE} & \multicolumn{1}{l}{RMSE} \\ \midrule
t+1  & 13.22  & 5.44  & 23.60  & 11.25  & 5.16  & 20.26  & 13.22  & 5.38  & 22.59  \\
t+2  & 13.81  & 5.82  & 24.65  & 14.08  & 6.37  & 24.35  & 15.24  & 6.20  & 26.05  \\
t+3  & 15.65  & 6.78  & 27.57  & 17.25  & 7.30  & 28.95  & 18.05  & 7.35  & 30.85  \\
t+4  & 18.90  & 7.99  & 32.69  & 21.96  & 8.84  & 36.92  & 23.50  & 9.56  & 40.15  \\
t+5  & 21.53  & 9.16  & 36.90  & 26.56  & 10.68 & 44.41  & 28.03  & 11.41 & 47.90  \\
t+6  & 24.82  & 10.43 & 42.42  & 31.05  & 12.08 & 51.93  & 32.57  & 13.26 & 55.66  \\
t+7  & 29.21  & 12.18 & 50.15  & 36.41  & 14.07 & 61.21  & 43.08  & 17.53 & 73.62  \\
t+8  & 34.73  & 14.12 & 60.09  & 44.45  & 17.45 & 73.72  & 55.72  & 22.68 & 95.22  \\
t+9  & 39.80  & 15.93 & 68.38  & 51.27  & 19.85 & 85.82  & 68.78  & 27.99 & 117.55 \\
t+10 & 44.65  & 17.59 & 76.58  & 55.87  & 21.43 & 93.43  & 75.32  & 30.66 & 128.72 \\
t+11 & 50.80  & 19.80 & 86.26  & 62.69  & 23.97 & 104.22 & 81.20  & 33.05 & 138.77 \\
t+12 & 54.66  & 21.32 & 92.63  & 67.87  & 25.94 & 112.90 & 87.11  & 35.45 & 148.86 \\
t+13 & 59.23  & 22.84 & 100.85 & 72.49  & 27.53 & 121.33 & 92.59  & 37.68 & 158.23 \\
t+14 & 64.76  & 24.95 & 109.85 & 79.52  & 30.26 & 133.18 & 113.36 & 46.14 & 193.73 \\
t+15 & 70.58  & 28.31 & 119.30 & 84.63  & 32.85 & 143.34 & 135.01 & 54.95 & 230.73 \\
t+16 & 75.80  & 30.31 & 127.58 & 90.83  & 35.32 & 152.41 & 157.40 & 64.06 & 268.99 \\
t+17 & 81.04  & 31.97 & 136.60 & 97.02  & 37.82 & 162.65 & 164.53 & 66.96 & 281.17 \\
t+18 & 85.72  & 33.91 & 143.86 & 101.87 & 40.38 & 170.64 & 171.76 & 69.91 & 293.54 \\
t+19 & 89.83  & 35.51 & 150.32 & 106.62 & 42.51 & 178.60 & 178.76 & 72.75 & 305.50 \\
t+20 & 92.34  & 36.70 & 154.14 & 110.32 & 44.28 & 185.29 & 185.37 & 75.45 & 316.80 \\
t+21 & 96.89  & 39.10 & 162.60 & 115.22 & 47.40 & 193.89 & 199.40 & 81.15 & 340.76 \\
t+22 & 97.55  & 39.57 & 164.07 & 115.21 & 48.60 & 192.58 & 211.81 & 86.21 & 361.99 \\
t+23 & 102.09 & 41.72 & 172.69 & 120.11 & 51.60 & 201.04 & 224.17 & 91.24 & 383.11 \\
t+24 & 105.86 & 43.46 & 180.38 & 126.08 & 55.03 & 212.00 & 227.43 & 92.56 & 388.67 \\
t+25 & 109.86 & 45.47 & 187.45 & 128.76 & 56.98 & 217.33 & 233.82 & 95.17 & 399.60 \\
t+26 & 112.42 & 46.88 & 192.31 & 133.86 & 59.91 & 226.53 & 239.23 & 97.37 & 408.84 \\
t+27 & 115.48 & 48.65 & 198.44 & 137.75 & 62.69 & 233.29 & 243.24 & 99.00 & 415.70 \\
t+28 & 118.38 & 50.66 & 204.09 & 141.74 & 66.53 & 240.39 & 243.44 & 99.08 & 416.03 \\ \bottomrule
\end{tabular}}
\label{spatial_eval_table}
\end{table}

\begin{table}[ht]
\tbl{Average 28-day–ahead errors for state-level
COVID-19–hospitalization forecasts during the Delta (15 weekly forecasts, June 21 – September 27, 2021) wave.}
{\scriptsize
\begin{tabular}{l*{12}{S}}
\toprule
      & \multicolumn{3}{c}{\textbf{SLSTM (w/ SPH)}} &
        \multicolumn{3}{c}{\textbf{COVIDhub‑baseline}} &
        \multicolumn{3}{c}{\textbf{COVIDhub‑4\_wk\_ensemble}} &
        \multicolumn{3}{c}{\textbf{COVIDhub‑trained\_ensemble}}\\
\cmidrule(lr){2-4}\cmidrule(lr){5-7}\cmidrule(lr){8-10}\cmidrule(lr){11-13}
{Day} & {MAE} & {MAPE} & {RMSE}
      & {MAE} & {MAPE} & {RMSE}
      & {MAE} & {MAPE} & {RMSE}
      & {MAE} & {MAPE} & {RMSE}\\
\midrule
t+1  &  5.35 &  5.71 &  10.65 & 15.31 & 28.93 & 31.25 & 15.17 & 15.22 & 32.52 & 19.24 & 17.63 & 39.45 \\
t+2  &  7.61 &  7.87 &  15.18 & 17.73 & 29.67 & 35.85 & 17.07 & 16.41 & 35.88 & 21.92 & 19.41 & 45.59 \\
t+3  &  9.84 &  9.76 &  19.76 & 20.78 & 31.08 & 41.70 & 20.09 & 17.86 & 42.62 & 25.28 & 21.11 & 53.22 \\
t+4  & 12.07 & 11.50 &  24.53 & 23.82 & 30.79 & 47.94 & 22.81 & 19.79 & 50.07 & 27.66 & 22.22 & 58.71 \\
t+5  & 14.11 & 13.46 &  28.72 & 26.53 & 32.09 & 53.58 & 23.84 & 21.82 & 50.88 & 28.51 & 24.02 & 61.27 \\
t+6  & 16.29 & 15.04 &  33.27 & 29.21 & 32.44 & 59.62 & 25.80 & 23.00 & 55.21 & 31.51 & 25.42 & 67.24 \\
t+7  & 18.55 & 16.44 &  38.05 & 32.45 & 32.56 & 66.51 & 30.40 & 24.08 & 63.83 & 38.39 & 27.71 & 80.22 \\
t+8  & 22.47 & 18.99 &  45.66 & 35.70 & 32.58 & 73.35 & 32.04 & 25.04 & 69.86 & 41.85 & 29.06 & 88.36 \\
t+9  & 24.83 & 19.98 &  50.67 & 39.00 & 33.07 & 80.57 & 35.09 & 26.77 & 77.00 & 45.15 & 31.23 & 94.81 \\
t+10 & 27.63 & 21.81 &  56.16 & 42.36 & 34.70 & 87.75 & 38.15 & 28.35 & 81.84 & 49.51 & 33.03 & 104.43 \\
t+11 & 29.66 & 23.01 &  60.45 & 45.69 & 35.64 & 94.96 & 41.28 & 29.46 & 90.08 & 52.53 & 34.64 & 111.59 \\
t+12 & 32.04 & 24.17 &  65.28 & 48.83 & 37.00 & 101.65& 41.98 & 31.15 & 90.17 & 53.73 & 35.77 & 114.62 \\
t+13 & 34.13 & 25.63 &  69.85 & 51.93 & 37.93 & 108.27& 44.48 & 32.57 & 95.44 & 57.71 & 37.46 & 122.53 \\
t+14 & 37.19 & 27.25 &  76.01 & 55.44 & 39.85 & 115.55& 50.41 & 34.24 & 104.60& 66.74 & 40.84 & 140.52 \\
t+15 & 47.76 & 34.73 &  96.57 & 58.83 & 41.67 & 122.75& 51.28 & 34.55 & 111.31& 70.24 & 41.99 & 148.63 \\
t+16 & 50.68 & 36.43 & 102.34 & 62.20 & 42.70 & 129.96& 53.18 & 36.13 & 112.51& 75.28 & 44.60 & 159.04 \\
t+17 & 53.54 & 38.03 & 107.78 & 65.46 & 44.07 & 137.05& 56.27 & 37.63 & 118.51& 80.25 & 46.35 & 169.88 \\
t+18 & 56.06 & 39.10 & 113.19 & 68.74 & 45.68 & 144.03& 59.96 & 38.80 & 128.76& 85.48 & 48.37 & 182.20 \\
t+19 & 58.88 & 40.50 & 118.84 & 71.81 & 46.98 & 150.39& 60.15 & 39.94 & 126.25& 86.77 & 49.64 & 183.81 \\
t+20 & 61.35 & 42.06 & 123.19 & 74.83 & 48.21 & 156.70& 63.56 & 41.46 & 134.35& 91.85 & 51.82 & 193.60 \\
t+21 & 64.22 & 43.08 & 128.57 & 78.17 & 49.67 & 163.53& 70.29 & 44.85 & 145.35& 104.44& 57.22 & 219.54 \\
t+22 & 70.40 & 47.39 & 139.56 & 81.55 & 51.09 & 170.46& 70.76 & 43.85 & 151.99& 108.53& 58.81 & 227.93 \\
t+23 & 73.33 & 48.90 & 145.01 & 84.82 & 52.61 & 177.17& 72.39 & 45.57 & 151.37& 114.15& 61.58 & 239.07 \\
t+24 & 76.40 & 50.13 & 150.40 & 87.99 & 53.81 & 183.74& 75.95 & 47.07 & 158.66& 120.81& 64.17 & 250.55 \\
t+25 & 79.12 & 51.34 & 155.29 & 91.07 & 55.17 & 190.11& 79.92 & 48.18 & 168.27& 125.73& 67.08 & 259.32 \\
t+26 & 81.25 & 52.21 & 159.66 & 93.81 & 56.37 & 195.81& 79.66 & 48.91 & 165.74& 126.04& 67.78 & 256.78 \\
t+27 & 83.47 & 53.12 & 164.11 & 96.58 & 57.70 & 201.40& 82.35 & 49.86 & 170.56& 131.90& 70.31 & 268.71 \\
t+28 & 86.38 & 54.10 & 169.62 & 99.58 & 59.27 & 207.40& 89.28 & 53.64 & 183.22& 147.34& 77.08 & 302.03 \\
\bottomrule
\end{tabular}}
\label{tbl:baseline_eval_combined_delta}
\end{table}

\begin{table}[ht]
\tbl{Average 28-day–ahead errors for state-level
COVID-19–hospitalization forecasts during the Omicron (10 weekly forecasts, December 6, 2021 – February 7, 2022) wave. }
{\scriptsize
\begin{tabular}{lcccccccccccc}
\toprule
& \multicolumn{3}{c}{SLSTM (w/ SPH)} &
  \multicolumn{3}{c}{COVIDhub-baseline} &
  \multicolumn{3}{c}{COVIDhub-4\_wk\_ensemble} &
  \multicolumn{3}{c}{COVIDhub-trained\_ensemble} \\ 
\cmidrule(lr){2-4}\cmidrule(lr){5-7}\cmidrule(lr){8-10}\cmidrule(lr){11-13}
Day
 & MAE & MAPE & RMSE
 & MAE & MAPE & RMSE
 & MAE & MAPE & RMSE
 & MAE & MAPE & RMSE \\ \midrule
t+1  & 13.22 &  5.44 &  23.60 &  28.17 & 12.84 &  52.91 &  22.05 &  9.15 &  38.96 &  24.38 &  9.64 &  43.94 \\
t+2  & 13.81 &  5.82 &  24.65 &  35.49 & 14.97 &  64.40 &  28.06 & 11.11 &  48.03 &  28.74 & 11.19 &  50.13 \\
t+3  & 15.65 &  6.78 &  27.57 &  44.05 & 17.55 &  78.18 &  34.18 & 13.17 &  57.69 &  34.96 & 13.29 &  60.36 \\
t+4  & 18.90 &  7.99 &  32.69 &  52.82 & 20.64 &  91.98 &  38.96 & 15.16 &  65.89 &  41.70 & 15.54 &  71.21 \\
t+5  & 21.53 &  9.16 &  36.90 &  60.09 & 22.97 & 104.01 &  43.05 & 16.56 &  71.44 &  47.13 & 17.33 &  81.93 \\
t+6  & 24.82 & 10.43 &  42.42 &  67.21 & 25.23 & 115.92 &  48.35 & 18.01 &  81.03 &  55.03 & 19.33 &  96.56 \\
t+7  & 29.21 & 12.18 &  50.15 &  75.68 & 28.07 & 129.99 &  56.83 & 20.73 &  97.23 &  67.50 & 22.55 & 118.06 \\
t+8  & 34.73 & 14.12 &  60.09 &  84.31 & 31.35 & 144.57 &  64.32 & 23.01 & 110.04 &  76.96 & 25.16 & 135.70 \\
t+9  & 39.80 & 15.93 &  68.38 &  92.77 & 34.53 & 159.02 &  70.17 & 24.79 & 121.44 &  86.65 & 27.50 & 152.87 \\
t+10 & 44.65 & 17.59 &  76.58 & 101.50 & 38.05 & 173.95 &  77.16 & 27.28 & 133.62 &  97.51 & 30.50 & 172.03 \\
t+11 & 50.80 & 19.80 &  86.26 & 109.92 & 41.94 & 188.04 &  84.61 & 29.95 & 146.08 & 106.70 & 33.57 & 188.72 \\
t+12 & 54.66 & 21.32 &  92.63 & 116.78 & 45.24 & 200.12 &  88.89 & 31.14 & 153.92 & 114.21 & 35.72 & 203.71 \\
t+13 & 59.23 & 22.84 & 100.85 & 123.40 & 48.48 & 211.95 &  94.53 & 32.97 & 163.89 & 121.94 & 38.12 & 216.97 \\
t+14 & 64.76 & 24.95 & 109.85 & 131.57 & 52.90 & 225.78 & 106.92 & 37.12 & 191.40 & 135.55 & 42.68 & 239.53 \\
t+15 & 70.58 & 28.31 & 119.30 & 139.95 & 57.08 & 239.78 & 115.66 & 40.23 & 209.51 & 147.00 & 46.54 & 263.36 \\
t+16 & 75.80 & 30.31 & 127.58 & 147.75 & 61.57 & 253.31 & 122.78 & 43.07 & 222.98 & 156.98 & 50.24 & 282.60 \\
t+17 & 81.04 & 31.97 & 136.60 & 155.27 & 66.04 & 266.74 & 130.04 & 45.79 & 237.04 & 166.09 & 53.46 & 300.55 \\
t+18 & 85.72 & 33.91 & 143.86 & 162.51 & 71.06 & 279.31 & 136.68 & 48.91 & 249.34 & 173.82 & 56.69 & 316.66 \\
t+19 & 89.83 & 35.51 & 150.32 & 168.88 & 75.31 & 290.22 & 142.07 & 50.63 & 259.80 & 177.58 & 58.84 & 328.17 \\
t+20 & 92.34 & 36.70 & 154.14 & 174.89 & 79.75 & 300.57 & 145.83 & 52.67 & 265.70 & 183.11 & 61.53 & 337.69 \\
t+21 & 96.89 & 39.10 & 162.60 & 181.98 & 85.69 & 312.74 & 151.66 & 57.67 & 273.62 & 194.28 & 67.05 & 360.11 \\
t+22 & 97.55 & 39.57 & 164.07 & 189.21 & 91.67 & 324.62 & 160.89 & 62.42 & 290.96 & 201.72 & 72.05 & 375.70 \\
t+23 & 102.09 & 41.72 & 172.69 & 196.05 & 98.10 & 335.29 & 166.98 & 66.11 & 302.62 & 207.15 & 75.82 & 387.08 \\
t+24 & 105.86 & 43.46 & 180.38 & 202.59 & 104.74 & 345.82 & 171.72 & 69.47 & 310.78 & 210.50 & 79.37 & 391.90 \\
t+25 & 109.86 & 45.47 & 187.45 & 208.61 & 111.15 & 355.61 & 176.65 & 73.21 & 319.28 & 215.25 & 84.19 & 402.53 \\
t+26 & 112.42 & 46.88 & 192.31 & 213.22 & 117.01 & 363.32 & 178.38 & 74.47 & 323.35 & 217.12 & 86.68 & 410.14 \\
t+27 & 115.48 & 48.65 & 198.44 & 217.32 & 122.77 & 370.23 & 181.42 & 76.96 & 328.62 & 218.45 & 93.11 & 401.18 \\
t+28 & 118.38 & 50.66 & 204.09 & 221.85 & 130.54 & 377.99 & 187.42 & 83.71 & 341.58 & 225.43 &100.13 & 413.37 \\
\bottomrule
\end{tabular}}
\label{tbl:baseline_eval_combined_omicron}
\end{table}

\begin{table}[ht]
\tbl{Decomposition of the weighted‑interval score (WIS) by forecast horizon for the
Delta wave (15 weekly forecasts, June 21 – September 27, 2021):   
Dispersion, under-prediction penalty, over-prediction penalty, and total WIS
(Lower is better).}
{\scriptsize
\begin{tabular}{l*{16}{S}}
\toprule
      & \multicolumn{4}{c}{\textbf{SLSTM (w/ SPH)}} &
        \multicolumn{4}{c}{\textbf{COVIDhub‑baseline}} &
        \multicolumn{4}{c}{\textbf{COVIDhub‑4wk ensemble}} &
        \multicolumn{4}{c}{\textbf{COVIDhub‑trained ensemble}} \\
\cmidrule(lr){2-5}\cmidrule(lr){6-9}\cmidrule(lr){10-13}\cmidrule(lr){14-17}
{Day} &
\multicolumn{1}{c}{Disp} & \multicolumn{1}{c}{Under} & \multicolumn{1}{c}{Over} & \multicolumn{1}{c}{WIS} &
\multicolumn{1}{c}{Disp} & \multicolumn{1}{c}{Under} & \multicolumn{1}{c}{Over} & \multicolumn{1}{c}{WIS} &
\multicolumn{1}{c}{Disp} & \multicolumn{1}{c}{Under} & \multicolumn{1}{c}{Over} & \multicolumn{1}{c}{WIS} &
\multicolumn{1}{c}{Disp} & \multicolumn{1}{c}{Under} & \multicolumn{1}{c}{Over} & \multicolumn{1}{c}{WIS} \\
\midrule
t+1  & 1.54 & 0.60 & 1.35 & 3.50 & 8.16 & 1.70 & 2.90 & 12.76 & 8.08 & 1.57 & 1.70 & 11.36 & 9.27 & 1.43 & 3.28 & 13.99 \\
t+2  & 1.96 & 1.05 & 1.90 & 4.91 & 9.35 & 2.10 & 2.93 & 14.38 & 8.39 & 1.78 & 2.27 & 12.44 & 9.60 & 1.62 & 4.21 & 15.44 \\
t+3  & 2.37 & 1.54 & 2.39 & 6.29 & 10.39 & 2.68 & 3.14 & 16.21 & 8.81 & 2.11 & 2.97 & 13.89 & 10.15 & 1.98 & 5.00 & 17.13 \\
t+4  & 2.82 & 2.03 & 2.88 & 7.72 & 11.33 & 3.33 & 3.42 & 18.08 & 9.08 & 2.65 & 3.55 & 15.28 & 10.45 & 2.40 & 5.75 & 18.60 \\
t+5  & 3.30 & 2.41 & 3.30 & 9.01 & 12.18 & 3.95 & 3.64 & 19.77 & 9.07 & 3.74 & 3.04 & 15.85 & 10.61 & 3.38 & 5.47 & 19.46 \\
t+6  & 3.78 & 2.87 & 3.70 & 10.35 & 12.97 & 4.59 & 3.94 & 21.50 & 9.26 & 4.57 & 3.47 & 17.30 & 10.90 & 3.96 & 6.53 & 21.39 \\
t+7  & 4.21 & 3.47 & 4.15 & 11.83 & 13.70 & 5.40 & 4.30 & 23.39 & 10.14 & 4.00 & 5.52 & 19.66 & 11.84 & 3.62 & 9.46 & 24.92 \\
t+8  & 4.75 & 4.41 & 5.11 & 14.27 & 14.39 & 6.22 & 4.70 & 25.30 & 10.30 & 4.79 & 5.94 & 21.03 & 12.47 & 4.08 & 10.43 & 26.98 \\
t+9  & 5.21 & 5.05 & 5.56 & 15.82 & 15.03 & 7.12 & 5.14 & 27.29 & 10.60 & 5.58 & 6.51 & 22.69 & 12.85 & 4.39 & 11.84 & 29.08 \\
t+10 & 5.59 & 5.77 & 6.09 & 17.45 & 15.65 & 8.02 & 5.61 & 29.28 & 11.16 & 6.26 & 7.21 & 24.63 & 13.38 & 4.96 & 13.45 & 31.79 \\
t+11 & 5.99 & 6.47 & 6.56 & 19.03 & 16.24 & 8.94 & 6.12 & 31.29 & 11.50 & 7.13 & 7.78 & 26.41 & 13.94 & 5.71 & 14.36 & 34.01 \\
t+12 & 6.38 & 7.13 & 7.04 & 20.55 & 16.80 & 9.75 & 6.60 & 33.15 & 11.36 & 8.56 & 7.03 & 26.94 & 13.98 & 6.88 & 14.22 & 35.07 \\
t+13 & 6.79 & 7.84 & 7.45 & 22.08 & 17.34 & 10.58 & 7.10 & 35.01 & 11.62 & 9.92 & 7.12 & 28.66 & 14.52 & 7.64 & 15.62 & 37.78 \\
t+14 & 7.20 & 8.76 & 7.92 & 23.88 & 17.86 & 11.56 & 7.61 & 37.03 & 13.06 & 9.08 & 9.81 & 31.95 & 15.93 & 7.27 & 19.98 & 43.18 \\
t+15 & 7.26 & 15.75 & 9.20 & 32.21 & 18.36 & 12.56 & 8.11 & 39.02 & 13.06 & 10.01 & 10.06 & 33.13 & 16.41 & 8.00 & 21.48 & 45.88 \\
t+16 & 7.62 & 16.78 & 9.74 & 34.13 & 18.84 & 13.55 & 8.69 & 41.07 & 13.43 & 11.05 & 9.84 & 34.33 & 17.14 & 8.81 & 22.98 & 48.93 \\
t+17 & 7.96 & 17.83 & 10.26 & 36.04 & 19.31 & 14.47 & 9.29 & 43.06 & 14.01 & 11.97 & 10.56 & 36.53 & 17.96 & 9.64 & 25.03 & 52.64 \\
t+18 & 8.24 & 18.77 & 10.80 & 37.81 & 19.76 & 15.41 & 9.83 & 45.01 & 14.30 & 13.13 & 11.39 & 38.82 & 18.45 & 10.72 & 26.76 & 55.93 \\
t+19 & 8.54 & 19.73 & 11.31 & 39.58 & 20.20 & 16.28 & 10.32 & 46.80 & 14.22 & 15.25 & 9.88 & 39.35 & 18.67 & 12.21 & 26.56 & 57.43 \\
t+20 & 8.84 & 20.59 & 11.84 & 41.28 & 20.63 & 17.08 & 10.80 & 48.51 & 14.39 & 16.96 & 10.22 & 41.57 & 19.40 & 13.23 & 28.55 & 61.18 \\
t+21 & 9.19 & 21.61 & 12.30 & 43.10 & 21.04 & 17.97 & 11.49 & 50.51 & 15.76 & 16.31 & 13.69 & 45.77 & 21.03 & 13.12 & 34.90 & 69.04 \\
t+22 & 9.54 & 24.61 & 13.29 & 47.45 & 21.45 & 18.94 & 12.12 & 52.51 & 15.78 & 17.57 & 13.41 & 46.76 & 21.60 & 14.21 & 36.80 & 72.60 \\
t+23 & 9.85 & 25.71 & 13.91 & 49.47 & 21.85 & 19.87 & 12.72 & 54.44 & 15.94 & 18.67 & 13.24 & 47.85 & 22.52 & 15.16 & 38.75 & 76.43 \\
t+24 & 10.13 & 26.55 & 14.47 & 51.15 & 22.24 & 20.81 & 13.27 & 56.32 & 16.47 & 20.02 & 13.78 & 50.27 & 23.34 & 16.54 & 40.76 & 80.64 \\
t+25 & 10.45 & 27.47 & 14.88 & 52.80 & 22.61 & 21.68 & 13.87 & 58.16 & 16.76 & 21.37 & 14.43 & 52.56 & 24.14 & 17.70 & 42.34 & 84.18 \\
t+26 & 10.69 & 28.18 & 15.48 & 54.35 & 22.99 & 22.42 & 14.38 & 59.79 & 16.52 & 23.75 & 12.86 & 53.13 & 24.15 & 19.45 & 41.63 & 85.22 \\
t+27 & 11.04 & 28.96 & 15.79 & 55.79 & 23.35 & 23.15 & 14.91 & 61.41 & 16.70 & 25.40 & 13.14 & 55.24 & 24.69 & 20.68 & 43.99 & 89.36 \\
t+28 & 11.35 & 29.83 & 16.38 & 57.56 & 23.71 & 23.96 & 15.46 & 63.12 & 17.81 & 24.99 & 16.59 & 59.39 & 26.51 & 20.67 & 51.94 & 99.11 \\
\bottomrule
\end{tabular}}
\label{tab:wis_decomposition_delta}
\end{table}

\begin{table}[ht]
\tbl{Decomposition of the weighted-interval score (WIS) by forecast horizon during the Omicron wave (10 weekly forecasts, December 6, 2021 – February 7, 2022):
Dispersion, under-prediction penalty, over-prediction penalty, and total WIS
(Lower is better).}
{\scriptsize
\begin{tabular}{l*{16}{S}}        
\toprule
      & \multicolumn{4}{c}{SLSTM (w/ SPH)} %
      & \multicolumn{4}{c}{COVIDhub-baseline} %
      & \multicolumn{4}{c}{COVIDhub-4\_wk\_ensemble} %
      & \multicolumn{4}{c}{COVIDhub-trained\_ensemble} \\
\cmidrule(lr){2-5}\cmidrule(lr){6-9}\cmidrule(lr){10-13}\cmidrule(lr){14-17}
{Day} &
\multicolumn{1}{c}{Disp} & \multicolumn{1}{c}{Under} & \multicolumn{1}{c}{Over} & \multicolumn{1}{c}{WIS} &
\multicolumn{1}{c}{Disp} & \multicolumn{1}{c}{Under} & \multicolumn{1}{c}{Over} & \multicolumn{1}{c}{WIS} &
\multicolumn{1}{c}{Disp} & \multicolumn{1}{c}{Under} & \multicolumn{1}{c}{Over} & \multicolumn{1}{c}{WIS} &
\multicolumn{1}{c}{Disp} & \multicolumn{1}{c}{Under} & \multicolumn{1}{c}{Over} & \multicolumn{1}{c}{WIS} \\
\midrule
t+1  & 4.1999 & 3.9068 & 0.9208 & 9.0275
     & 10.8709 & 4.1102 & 5.9875 & 19.3622
     & 9.6427 & 2.3591 & 3.7894 & 15.3634
     & 9.1835 & 4.3347 & 2.9486 & 16.2655 \\
t+2  & 4.3816 & 4.1051 & 0.8444 & 9.3311
     & 12.4659 & 6.0250 & 7.1892 & 23.7256
     & 10.5692 & 3.5321 & 4.9660 & 18.5750
     & 9.7767 & 4.7066 & 4.2715 & 18.4912 \\
t+3  & 4.7910 & 4.8320 & 0.9839 & 10.6068
     & 13.8744 & 8.4644 & 8.9945 & 29.0759
     & 11.3539 & 5.1627 & 6.3831 & 22.4386
     & 10.6099 & 5.7749 & 6.6493 & 22.7701 \\
t+4  & 5.2585 & 5.9600 & 1.3826 & 12.6010
     & 15.2150 & 11.0734 & 10.9368 & 34.6380
     & 12.1624 & 6.9470 & 7.2352 & 25.9107
     & 11.5733 & 6.9893 & 9.2639 & 27.5004 \\
t+5  & 5.6915 & 7.0657 & 1.7811 & 14.5384
     & 16.5091 & 13.2172 & 12.6141 & 39.3916
     & 12.6150 & 9.3956 & 6.8769 & 28.5228
     & 12.2674 & 9.3057 & 10.3693 & 31.5694 \\
t+6  & 6.2331 & 7.9457 & 2.2255 & 16.4043
     & 17.7749 & 15.2469 & 14.3625 & 44.0471
     & 13.1176 & 10.9755 & 8.1065 & 31.8575
     & 13.2003 & 11.2131 & 13.0331 & 37.0408 \\
t+7  & 6.8349 & 9.4713 & 2.9898 & 19.2960
     & 19.1130 & 17.6533 & 16.6695 & 49.5779
     & 14.0335 & 10.7943 & 12.4193 & 36.8035
     & 14.2828 & 11.0189 & 20.3190 & 45.1939 \\
t+8  & 7.8382 & 11.3024 & 3.5063 & 22.6469
     & 20.4071 & 20.1889 & 19.2142 & 55.4331
     & 15.3097 & 12.3912 & 14.6354 & 41.8428
     & 15.4716 & 12.6810 & 23.5540 & 51.2480 \\
t+9  & 8.3594 & 13.2204 & 4.3232 & 25.9030
     & 21.6392 & 22.7911 & 21.7223 & 61.2805
     & 15.9194 & 14.4492 & 16.2097 & 46.0936
     & 16.4637 & 14.8739 & 27.2712 & 58.1535 \\
t+10 & 8.8713 & 15.3734 & 5.2086 & 29.4534
     & 22.8223 & 25.5465 & 24.4000 & 67.4140
     & 16.6418 & 16.5644 & 17.8776 & 50.5617
     & 17.4262 & 17.2480 & 30.6824 & 64.8616 \\
t+11 & 9.3528 & 17.6513 & 6.2400 & 33.2441
     & 23.9768 & 28.1945 & 27.0199 & 73.3589
     & 17.2160 & 18.9621 & 19.3815 & 55.0636
     & 18.2774 & 19.8437 & 34.1136 & 71.7252 \\
t+12 & 9.8987 & 19.3790 & 6.8606 & 36.1383
     & 24.9916 & 30.1808 & 29.2266 & 78.2050
     & 17.7211 & 21.7865 & 18.9527 & 58.0505
     & 19.2267 & 23.4302 & 34.0285 & 76.1552 \\
t+13 & 10.4209 & 21.2143 & 7.7507 & 39.3858
     & 25.9162 & 32.0356 & 31.3638 & 82.8260
     & 18.1969 & 23.7250 & 20.5304 & 62.0454
     & 20.2141 & 25.9079 & 36.1789 & 81.7486 \\
t+14 & 10.8845 & 23.5778 & 8.7823 & 43.2445
     & 26.8767 & 34.3424 & 34.0591 & 88.4285
     & 19.0434 & 23.0867 & 27.7564 & 69.3795
     & 20.6253 & 25.2365 & 46.6575 & 91.9207 \\
t+15 & 10.8215 & 24.7231 & 12.6078 & 48.1523
     & 27.7950 & 36.7191 & 36.8560 & 94.1862
     & 19.9439 & 24.4947 & 31.0077 & 74.9013
     & 21.4837 & 26.8709 & 51.0305 & 98.7876 \\
t+16 & 11.3241 & 26.7934 & 13.6667 & 51.7842
     & 28.6696 & 38.9938 & 39.5075 & 99.6834
     & 20.6341 & 26.3636 & 33.2506 & 79.7081
     & 22.0436 & 29.4415 & 54.7448 & 105.6159 \\
t+17 & 11.8052 & 29.1635 & 14.5955 & 55.5642
     & 29.4837 & 41.3825 & 42.1382 & 105.2565
     & 21.1384 & 28.3440 & 35.6736 & 84.6227
     & 22.5859 & 31.9102 & 57.8115 & 111.6829 \\
t+18 & 12.3762 & 31.6261 & 15.4349 & 59.4373
     & 30.2221 & 43.6901 & 44.6566 & 110.6210
     & 21.6547 & 30.4620 & 37.3914 & 88.9959
     & 23.2828 & 34.5039 & 60.1326 & 117.2756 \\
t+19 & 12.8912 & 33.3361 & 16.0470 & 62.2743
     & 30.8872 & 45.2565 & 46.8188 & 114.8704
     & 21.9438 & 33.2464 & 37.6222 & 92.2616
     & 23.7022 & 38.0630 & 59.0490 & 120.1289 \\
t+20 & 13.4343 & 34.6763 & 16.3319 & 64.4425
     & 31.5368 & 46.5497 & 48.9406 & 118.7917
     & 22.4936 & 34.6976 & 38.5254 & 95.1126
     & 24.4587 & 40.2971 & 61.2406 & 125.2704 \\
t+21 & 13.9351 & 36.4179 & 17.3335 & 67.6866
     & 32.2249 & 48.2390 & 51.5970 & 123.6359
     & 23.3665 & 34.1013 & 43.3139 & 100.1555
     & 25.0840 & 38.7314 & 69.8066 & 132.8582 \\
t+22 & 13.7223 & 39.0656 & 16.0166 & 68.8044
     & 32.8877 & 49.9565 & 54.1905 & 128.4304
     & 23.9571 & 35.7570 & 47.4409 & 106.4522
     & 25.8057 & 40.7512 & 73.2348 & 138.9765 \\
t+23 & 14.1070 & 40.6039 & 16.8293 & 71.5402
     & 33.5223 & 51.4225 & 56.6847 & 132.8671
     & 24.5310 & 37.2763 & 49.3049 & 110.4213
     & 26.3397 & 43.1199 & 75.1637 & 143.8392 \\
t+24 & 14.5349 & 42.2689 & 17.6342 & 74.4380
     & 34.1355 & 52.8839 & 59.0604 & 137.1633
     & 24.8347 & 39.1492 & 50.9959 & 114.3276
     & 26.6599 & 45.2147 & 76.3082 & 147.4371 \\
t+25 & 14.9536 & 43.8572 & 18.3310 & 77.1418
     & 34.7173 & 54.3153 & 61.3699 & 141.3595
     & 25.1712 & 40.7497 & 52.9019 & 118.2031
     & 26.8889 & 47.1981 & 77.2227 & 150.5909 \\
t+26 & 15.3662 & 44.7108 & 18.8975 & 78.9745
     & 35.2404 & 55.0590 & 63.2973 & 144.4722
     & 25.1622 & 42.8181 & 51.9338 & 119.3020
     & 26.9479 & 49.9292 & 74.8379 & 151.0169 \\
t+27 & 15.7124 & 45.3643 & 19.4244 & 80.5012
     & 35.7553 & 55.4819 & 65.1743 & 147.2107
     & 25.4417 & 43.9100 & 52.9103 & 121.6639
     & 27.2704 & 51.0477 & 74.4205 & 151.9407 \\
t+28 & 16.1868 & 46.1203 & 20.2369 & 82.5440
     & 36.3280 & 55.9257 & 67.5995 & 150.5083
     & 25.6425 & 42.6253 & 58.8907 & 126.5463
     & 27.6665 & 49.1689 & 79.4008 & 155.4528 \\
\bottomrule
\end{tabular}}
\label{tab:wis_decomposition_omicron}
\end{table}

\end{document}